%% file: ms.tex
\documentclass[preprint]{article}

\usepackage{microtype}
\usepackage{graphicx}
\usepackage{subfigure}
\usepackage{booktabs} 

\usepackage{hyperref}
\usepackage{xspace}

\usepackage[accepted]{icml2024}

\usepackage{amsmath}
\usepackage{amssymb}
\usepackage{mathtools}
\usepackage{amsthm}
\usepackage{inconsolata}
\usepackage{enumitem}
\usepackage{nolbreaks}
\usepackage[frozencache,cachedir=mintedcache]{minted}

\usepackage[capitalize,noabbrev]{cleveref}

\theoremstyle{plain}

\theoremstyle{definition}

\theoremstyle{remark}

\usepackage[textsize=tiny]{todonotes}

\icmltitlerunning{Evaluation of LLMs on Syntax-Aware Code Fill-in-the-Middle Tasks}

\newcommand{\ours}{\nolbreaks{SAFIM}\xspace}

\begin{document}

\twocolumn[
\icmltitle{Evaluation of LLMs on Syntax-Aware Code Fill-in-the-Middle Tasks}



\icmlsetsymbol{equal}{*}

\begin{icmlauthorlist}
\icmlauthor{Linyuan Gong}{ucb}
\icmlauthor{Sida Wang}{meta}
\icmlauthor{Mostafa Elhoushi}{meta}
\icmlauthor{Alvin Cheung}{ucb}
\end{icmlauthorlist}

\icmlaffiliation{ucb}{University of California at Berkeley}
\icmlaffiliation{meta}{AI at Meta}

\icmlcorrespondingauthor{Linyuan Gong}{gly@berkeley.edu}

\icmlkeywords{Large Language Model, Fill-in-the-Middle, Code Generation, Code Completion, Model Evaluation}

\vskip 0.3in
]



\printAffiliationsAndNotice{}  

\input{contents/abstract}

\input{contents/introduction}

\input{contents/related_work}

\input{contents/benchmark_construction}

\input{contents/prompts_and_post_processing}

\input{contents/experimental_setup}

\input{contents/experimental_results}

\input{contents/conclusion}

\input{contents/acknowledgements}

\input{contents/broader_impact}

\bibliography{references}
\bibliographystyle{icml2024}

\newpage
\appendix
\onecolumn
\section{Appendix}

\input{contents/appendix/benchmark_api_libraries}

\input{contents/appendix/benchmark_statistics}

\input{contents/appendix/model_names}

\input{contents/appendix/all_results}

\input{contents/appendix/extra_results_truncation}

\input{contents/appendix/additional_main_results}

\input{contents/appendix/per_pl_results}

\input{contents/appendix/case_study}

\input{contents/appendix/data_contamination}


\end{document}

%% file: contents/abstract.tex
\begin{abstract}
We introduce \textbf{S}yntax-\textbf{A}ware \textbf{F}ill-\textbf{i}n-the-\textbf{M}iddle (\ours), a new benchmark for evaluating Large Language Models (LLMs) on the code Fill-in-the-Middle (FIM) task. This benchmark focuses on syntax-aware completions of program structures such as code blocks and conditional expressions, and includes 17,720 examples from multiple programming languages, sourced from recent code submissions after April 2022 to minimize data contamination. SAFIM provides a robust framework with various prompt designs and novel syntax-aware post-processing techniques, facilitating accurate and fair comparisons across LLMs. Our comprehensive evaluation of 15 LLMs shows that FIM pretraining not only enhances FIM proficiency but also improves Left-to-Right (L2R) inference using LLMs. Our findings challenge conventional beliefs and suggest that pretraining methods and data quality have more impact than model size. \ours thus serves as a foundational platform for future research in effective pretraining strategies for code LLMs.
The evaluation toolkit and dataset are available at \url{https://github.com/gonglinyuan/safim}, and the leaderboard is available at \url{https://safimbenchmark.com}.
\end{abstract}

%% file: contents/introduction.tex
\section{Introduction}\label{sec:introduction}

Recent advances in Large Language Models (LLMs) such as GPT-3.5~\citep{instructgpt}, GPT-4~\citep{gpt4}, and CodeLLaMa~\citep{codellama} have revolutionized coding-related tasks. However, existing benchmarks like HumanEval~\citep{codex} and MBPP~\citep{mbpp} focus on generating standalone functions or single-file code from natural language descriptions, and do not consider the more common practice of modifying and expanding existing code during development.

Recognizing this gap, we introduce the \textbf{S}yntax-\textbf{A}ware \textbf{F}ill-\textbf{i}n-the-\textbf{M}iddle (\ours) benchmark. \ours emphasizes syntax-aware completion within code's Abstract Syntax Tree (AST), targeting algorithmic blocks, control-flow expressions, and API function calls, unlike existing Fill-in-the Middle (FIM) benchmarks such as HumanEval-Infilling~\citep{humaneval_fim}, which are based on filling randomly masked lines or character spans. \ours is sourced from code on Codeforces and GitHub created after April 2022, deliberately aiming to avoid overlap with mainstream open-source pretraining corpora like The Stack~\citep{thestack}. This approach reduces the risks of data contamination caused by memoization of test cases, thereby bolstering the credibility of our results. \ours, with its 17,720 examples from 8,590 code files, not only surpasses the scale of HumanEval-Infilling, which draws from 164 short code files, but also expands the scope to include multiple programming languages. \ours primarily relies on execution-based evaluation, and uses syntactical match evaluation only when execution is not feasible due to external API calls.

Our comprehensive evaluation of 15 LLMs on \ours reveals its effectiveness in providing a fair comparison of models. We implement five distinct prompt designs to accommodate various model types and introduce a syntax-aware truncation algorithm for post-processing the outputs. Our approach unveils the true capabilities of non-FIM-trained models, allowing for a fair comparison with FIM-trained models.

Moreover, \ours sheds light on the strengths of various pretraining paradigms and challenges some prevalent beliefs in the field. Specifically, our findings suggest that FIM pretraining not only improves LLMs' performance in FIM inference but also enhances their performance in classical Left-to-Right (L2R) inference scenarios. This supports the growing trend of using FIM as the primary pretraining objective in code LLM development. 
We also observe that pretraining methods and data quality often outweigh the sheer model size---smaller models with sophisticated pretraining paradigms often outperform larger models. This is particularly evident in task-specific performances on \ours, where models pretrained with additional repo-level information excel in API function call completion, while those trained with code execution feedback perform better in control-flow expression generation. However, it is crucial to note that these comparisons across different model families are not controlled experiments and could be influenced by differences in pretraining environments. This suggests future work in pretraining such models under the same environment to validate these observations further. That said, our benchmark, \ours, provides a solid foundation for such future research, and opens up new opportunities in designing effective pretraining and fine-tuning paradigms for code LLMs.



%% file: contents/related_work.tex
\section{Related Work}\label{sec:related_work}

\paragraph{Large Language Models for Code.} The emergence of Large Language Models (LLMs) like GPT-3~\citep{gpt3} in natural language processing has led to the understanding that merely increasing the number of parameters in pretrained language models will ensure superior performance on unseen tasks. This has led to the application of LLMs to code-related tasks, particularly in code generation. For such tasks, decoder-only models are typically used. Initially, these models, such as Codex~\citep{codex}, PaLM~\citep{palm}, PolyCoder~\citep{polycoder}, and CodeGen~\citep{codegen}, primarily focused on Left-to-Right (L2R) pretraining, a.k.a. ``Next Token Prediction.'' However, the Fill-in-the-Middle (FIM) objective, a.k.a. ``Infilling,'' has become increasingly popular, with models like InCoder~\citep{incoder}, StarCoder~\citep{starcoder}, SantaCoder~\citep{santacoder}, DeepSeek-Coder~\citep{deepseekcoder}, and CodeLLaMa~\citep{codellama} showing their effectiveness. Additionally, proprietary models such as GPT-3.5~\citep{instructgpt}, GPT-4~\citep{gpt4}, and Gemini~\citep{gemini}, which use undisclosed pretraining methods, also contribute to this domain. While GLM-like models~\citep{glm} or encoder-decoder models, including CodeGeeX~\citep{codegeex}, PLBART~\citep{plbart}, AlphaCode~\citep{alphacode}, CodeT5~\citep{codet5, codet5p}, and AST-T5~\citep{astt5} exist, they are outside of our paper's scope. Our paper evaluates a select group of these LMs using the \ours benchmark. We develop insights into their performance in code FIM tasks, explore the strengths and weaknesses of various pretraining paradigms, and challenge the prevailing belief that a larger number of parameters automatically leads to better performance.

\vspace{-0.2in}

\paragraph{Benchmarking Generative Code LLMs.} Existing benchmarks for code generation in LLMs have a gap in effectively evaluating code generation capability for real-world development. Widely-used benchmarks like HumanEval~\citep{codex} and MBPP~\citep{mbpp} are limited to single Python functions and also subject to data contamination~\citep{datacontamination}. Extensions like HumanEval-X~\citep{codegeex}, MultiPLe~\citep{multiple}, and MBXP~\citep{mbxp} expand these benchmarks to other programming languages. Competition-style coding benchmarks like APPS~\citep{apps} and CodeContests~\citep{alphacode}, broaden the scope to file-level code generation. However, they still do not reflect typical development, which often involves iterative codebase expansion and invoking external API libraries. On the other hand, contextually richer benchmarks, such as JuICe~\citep{juice}, DS-1000~\citep{ds1000}, ARCADE~\citep{arcade}, NumpyEval~\citep{numpyeval}, and PandasEval~\citep{pandaseval}, PlotCoder~\citep{plotcoder}, ADELT~\citep{adelt} in data science, and APIBench~\citep{gorilla}, RepoBench~\citep{repobench}, ODEX~\citep{odex}, SWE-Bench~\citep{swebench}, GoogleCodeRepo~\citep{repolevelprompt}, RepoEval~\citep{repocoder}, and CoCoMIC-Data~\citep{cocomic} in software engineering, are often very small, heavily reliant on imperfect match-based evaluation metrics, or lacking in execution-based evaluation. Our \ours benchmark, based on Fill-in-the-Middle (FIM) tasks, bridges this gap by providing a comprehensive evaluation framework.

\paragraph{Fill-in-the-Middle in Training and Evaluating Code LLMs.} Fill-in-the-Middle (FIM) originates from masked language modeling (MLM) for training encoder-only models~\citep{bert} and T5-style span corruption for training encoder-decoder models~\citep{t5}, with span lengths usually limited to 1 to 5 tokens, with the goal of targeting representation learning rather than generation. For coding tasks, InCoder~\citep{incoder} shows the effectiveness of FIM as a pretraining objective for decoder-only models. \citet{incoder} further establishes the HumanEval-Infilling benchmark, further explored by \citet{humaneval_fim} in evaluating GPT-3/Codex variants, showing that a pretraining mix with a 90\% FIM ratio does not harm Left-to-Right (L2R) generation performance. CodeLLaMa's evaluations on HumanEval-Infilling support these findings, underscoring the value of FIM in pretraining code-focused LLMs~\citep{codellama}. However, this benchmark, limited to the 164 tiny Python snippets of HumanEval, emphasize the need for a more robust benchmark. \ours addresses this need by introducing a comprehensive, syntax-aware FIM benchmark for more detailed evaluations.

%% file: contents/benchmark_construction.tex
\section{Benchmark Construction}\label{sec:benchmark_construction}

The \ours benchmark is designed to evaluate Large Language Models (LLMs) on the Fill-in-the-Middle (FIM) of various code structures. In this section, we describe the collection of the corpora, the generation and filtering of completion tasks, and the evaluation protocols.

\input{figures/structfim_splits}

\subsection{Corpora Collection}

The \ours benchmark is constructed using corpora from two primary sources: \textit{Codeforces} and \textit{GitHub}. Codeforces,\footnote{\url{https://codeforces.com/}} a competitive programming platform, offers a wealth of coding problems, unit tests, and solutions. From Codeforces, we scrape problems, unit tests, and their corresponding code solutions. For GitHub, we gather git commits from the GH Archive\footnote{\url{https://www.gharchive.org/}}. From both sources, we gather Python, Java, C++, and C\# code files created between April 1, 2022, and January 1, 2023. This selection criteria ensures the inclusion of recent code, avoiding overlap with major pretraining datasets like The Stack~\citep{thestack} (cutoff at March 31, 2022) and the training data for GPT-3.5/GPT-4 (cutoff at September 2021), thus reducing the risk of data contamination.

In processing Codeforces data, we reevaluate each code solution by executing unit tests. We retain only those solutions that consistently pass all unit tests within 50\% of the specified time limit, eliminating randomness and noise from external factors. We also filter out excessively lengthy (over twice the size of the shortest accepted solution) or near-duplicate solutions (exceeding a CodeBLEU~\citep{codebleu} score threshold of 0.9 against previously added code), resulting in a curated set of 490 coding questions and 8,590 unique code solutions.

For GitHub, we first establish a list of widely-used API libraries for each programming language, detailed in \Cref{sec:benchmark_api_libraries}. We then extract code files that invoke APIs from such repositories with more than 10 stars to prioritize high-quality code. Files lacking natural language comments or documentation are excluded to avoid unsolvable examples. After thorough filtering and deduplication, our final GitHub corpus consists of 11,936 code files.

\subsection{Generating and Filtering Completion Tasks}

With our corpora ready, we parse each code file into an Abstract Syntax Tree (AST). This enables the creation of structured FIM tasks across three splits: algorithmic block completion, control-flow completion, and API function call completion. The first two are based on the Codeforces corpus, while the latter is based on the GitHub corpus as external API function calls are usually absent in competitive programming. In each split, we mask different code segments and ask the models to reconstruct these segments such that the original program functionality is maintained.

\paragraph{Algorithmic Block Completion.} Here, we mask a code block critical for solving the coding question, evaluating the LLM's capability in interpreting natural language descriptions and designing algorithms.  A ``code block'' refers to a contiguous list of statements, identified by indentations for Python or curly braces for C-family languages. We target the deepest block in the AST, often the innermost loop layer containing key operations or formulae, like a dynamic programming state transition equation (see \Cref{fig:structfim_splits}, Left). To avoid masking non-critical blocks (e.g., logging or debugging), we validate each block: if replacing a block with no-op causes unit test failures, it is included; otherwise, it is excluded. Such filtering ensures that only algorithmically significant blocks are included in the benchmark.

\paragraph{Control-Flow Completion.} This category focuses on masking critical control expressions in the program, evaluating the LLM's understanding of code control flows. We mask conditional expressions in statements such as \texttt{for}, \texttt{while}, \texttt{do-while}, \texttt{for-each}, \texttt{if}, and \texttt{else-if}. For example, in \Cref{fig:structfim_splits} (Middle), we mask \texttt{b \% 2} in an \texttt{if} statement, as it determines when the \texttt{result} variable will be updated; we mask \texttt{b > 0} of the outer layer \texttt{if} in a different example. To ensure the relevance of each masked expression, we only retain cases where substituting the expression with \texttt{false}, \texttt{true}, or an empty iterable would affect the unit test outcomes. Such filtering guarantees that only expressions critical to the program's control-flow are included in the benchmark. 

\paragraph{API Function Call Completion.} In this category, we mask calls to functions and object constructors from popular API libraries. This tests the LLM's API knowledge and the ability to integrate such knowledge with code context. Because this split is sourced from the inherently noisy GitHub corpus, we curate the dataset and add necessary hints as comments near each API call, ensuring each example is solvable by humans based on the given context. For example, in \Cref{fig:structfim_splits} (Right), the LLM is expected to deduce the correct arguments \texttt{max\_len} and \texttt{d\_model} for a positional embedding layer defined by \texttt{nn.Embedding}.

The \ours benchmark has 17,720 examples across these three categories, with detailed statistics provided in \Cref{sec:benchmark_statistics}.

\subsection{Evaluation Protocols}

We evaluate completions generated by LLMs using \textit{execution-based testing} and \textit{syntactical matching}. The former applies to algorithmic block and control-flow completions, while the latter is used for API function call completion.

{\bf Execution-Based Evaluation} is applied to examples with unit tests, covering 98.25\% of our benchmark. A completion is considered correct if it passes all unit tests. We use the ExecEval framework~\citep{xcodeval} as our execution environment for this purpose.

{\bf Syntactical Match Evaluation} is used where unit tests are impractical, which happens in the API function call completion split. This arises due to the potential side effects or dependencies on external environments inherent in external API function calls, which is difficult to check using only unit tests. In such instances, we use syntax matching to evaluate the model's output, comparing it against the ground truth. For instance, outputs like \texttt{func(a, b=1, c=2)} are considered equivalent to \texttt{func(a, c=2, b=1)}, focusing on syntactical equivalence rather than exact matches.

Our large dataset size of 17,720 examples enables robust evaluations without the need for multiple generations and averaging, as seen in smaller datasets like HumanEval (164 programs). Therefore, we only generate one completion for each LLM on each example and report the percentage of first-attempt passes, i.e., \textit{Pass@1}, as our evaluation metric.

%% file: figures/structfim_splits.tex
\begin{figure*}[ht]
\vskip 0.2in
\begin{center}
\centerline{\includegraphics{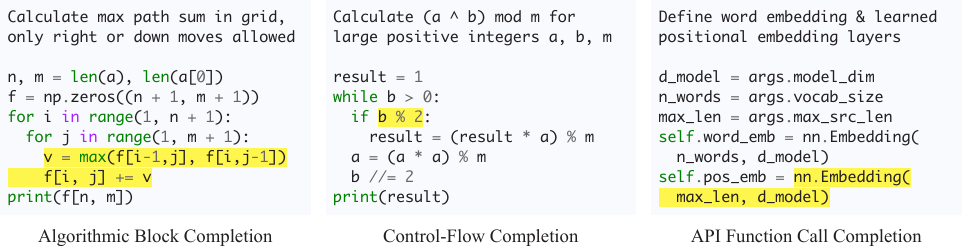}}
\caption{Three splits in the \ours benchmark illustrated with code examples. Each example includes a problem description and a code snippet, with a contiguous code segment highlighted in \colorbox{yellow}{yellow} to indicate the part to be masked and completed by LLMs. Contexts in these examples are shortened for clarity.}
\label{fig:structfim_splits}
\end{center}
\vskip -0.2in
\end{figure*}

%% file: contents/prompts_and_post_processing.tex
\section{Prompts and Post-Processing}\label{sec:prompts_and_post_processing}

We now describe our prompt designs and post-processing techniques for the \ours benchmark. These aspects make huge impact in model evaluations but are often overlooked. We introduce our approach for creating prompts and our unique syntax-aware post-processing method, which refines model outputs for more accurate and fair benchmarking.

\subsection{Prompts}\label{sec:prompts_and_post_processing_prompts}

\input{figures/structfim_prompts}

LLMs' performance is heavily influenced by the design of the prompts~\citep{prompt1,prompt2}. Using only a limited range of prompt types can skew evaluation results. For instance, \citet{incoder} use the Prefix-Suffix-Middle (PSM) prompt for FIM-pretrained models and the Instructed Prefix Feeding (IPF) prompt for others, leading to direct comparisons across different prompt types. This method, however, might yield suboptimal performance for different types of LLMs, leading to inaccurate comparisons. We further discuss this in \Cref{sec:experimental_results}. We address these concerns by introducing a wider range of distinct prompts in our evaluations, as detailed in \Cref{fig:structfim_prompts}:

\paragraph{Left-to-Right (L2R).} This baseline consists of only the code's prefix and omits the suffix. It provides a foundation to assess the effectiveness of other prompt designs.

\paragraph{Prefix-Suffix-Middle (PSM).} PSM uses a placeholder (a.k.a ``sentinel token'') to indicate the masked code segment, with the model tasked to generate the segment following the prompt. Effective use of this prompt type, however, requires that the model be pretrained with a FIM objective to recognize and appropriately respond to sentinel tokens.

\paragraph{Suffix-Prefix-Middle (SPM).} SPM places the suffix at the beginning and the completion segment immediately after the prefix. This structure enables models, even those not pretrained on FIM objectives like CodeGen, to perform the completion task in a left-to-right manner. This adaptability to non-FIM pretrained makes SPM suitable for a wider range of models, although \citet{codellama} reports SPM's inferior performance compared to PSM in the HumanEval-Infilling benchmark.

\paragraph{Instructed Prefix Feeding (IPF).} IPF replaces the masked code with a placeholder, followed by an instruction, and then repeats the prefix. It allows non-FIM pretrained models to recognize and tackle completion tasks~\citep{incoder}. Our experiments indicate a tendency in some models to erroneously output the placeholder token as part of their output. To address this, we introduce a logits masking technique to inhibit the generation of placeholder tokens, enhancing the effectiveness of IPF.

\paragraph{One-Shot (1S).} Tailored for non-FIM chat models, 1S uses a PSM-style prompt, supplemented with a simple input-output example, which provides the model with context about the task type and the expected input-output format.

\subsection{Post-Processing}\label{sec:prompts_and_post_processing_post_processing}

Post-processing is vital for automatic evaluation of LLMs in code generation, yet its importance is often underestimated. The raw output from LLMs is not immediately suitable for evaluation due to potential inclusions of irrelevant natural language or extra code beyond the targeted structure. \ours includes two stages of post-processing to address these challenges:

\paragraph{Code Extraction for Chat Models.} We use regex-based heuristics to extract code from outputs of chat models like GPT-4, which often mix natural language with code in the Markdown-formatted outputs.

\paragraph{Truncation.} An important challenge for models not fine-tuned for instruction following is their inability to determine the endpoint of their outputs. Often, such models generate the correct response but continue to produce extraneous content. A notable example is CodeGen~\citep{codegen}, which, due to its open-ended design, lacks the capability to signal an end-of-sequence (\texttt{<eos>}), resulting in unbounded output. Therefore, truncation is essential for the effective evaluation of code generation tasks.

However, inconsistencies in truncation methods across different models have led to skewed comparisons in prior work. For example, if the expected output is a Python expression and the truncation method retains only the first line of generated code, it may erroneously dismiss correct expressions that span multiple lines, as illustrated in \Cref{fig:structfim_splits} (Right).

\paragraph{Syntax-Aware Truncation.} In \ours, we introduce a syntax-aware truncation algorithm, replacing the conventional regex-based heuristics. This approach ensures the precise extraction of targeted code structures, thereby allowing for accurate and fair evaluations across different models.

For the algorithmic block completion task, which requires a code block as output, we use an iterative truncation process on the model's output. This involves sequentially removing the last line of the output until two key conditions are met: \textbf{(a)} the truncated output must fit into the AST as a ``code block'' subtree; and \textbf{(b)}, the AST of the remaining code---excluding the completion segment---must align with the AST of the original code, in terms of indentation level for Python or curly brace level for C-family languages. Once both conditions are satisfied, the truncated output is considered as the model's finalized completion.

For control-flow and API function call completions, our method incrementally adds characters to the output until it satisfies similar syntax matching criteria: the completed segment must form a valid ``expression'' node in the AST, and the rest of the code aligns precisely with the original code's AST structure.

%% file: figures/structfim_prompts.tex
\begin{figure*}[ht]
\vskip 0.2in
\begin{center}
\centerline{\includegraphics{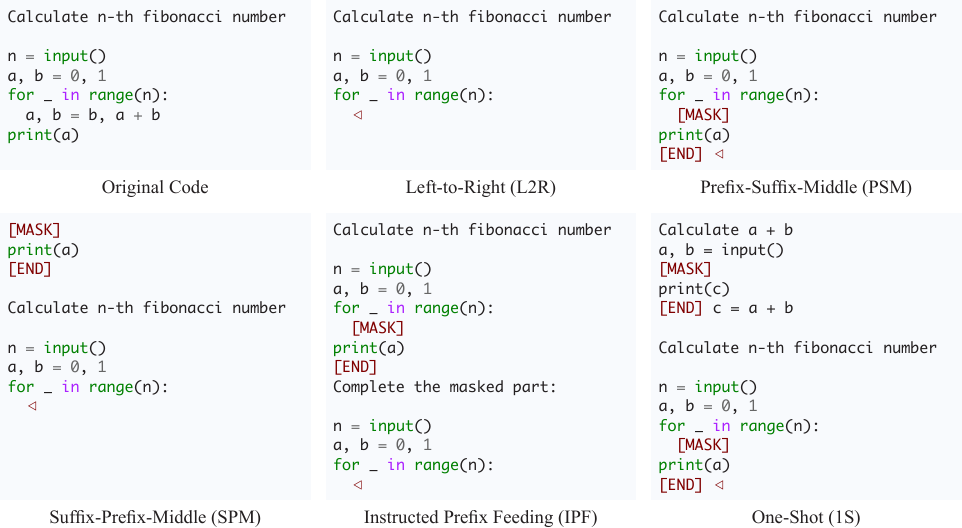}}
\caption{The original code is shown in the top-left, with the block \texttt{a, b = b, a + b} to be masked. The subsequent cells illustrate five distinct prompt types. The ``\textcolor[HTML]{800000}{\(\triangleleft\)}'' symbol indicates the end of the prompt, where model generation begins. The tokens \textcolor[HTML]{800000}{\texttt{[MASK]}} and \textcolor[HTML]{800000}{\texttt{[END]}} are model-specific, e.g., \textcolor[HTML]{800000}{\texttt{<SUF>}} and \textcolor[HTML]{800000}{\texttt{<MID>}} for CodeLLaMa, and \textcolor[HTML]{800000}{\texttt{<|mask:0|>}} and \textcolor[HTML]{800000}{\texttt{<|mask:1|>}} for InCoder.}
\label{fig:structfim_prompts}
\end{center}
\vspace{-0.2in}
\end{figure*}

%% file: contents/experimental_setup.tex
\section{Experimental Setup}\label{sec:experimental_setup}

\input{tables/table_models}

We evaluate GPT-3.5~\citep{gpt3,instructgpt}, GPT-4~\citep{gpt4}, CodeGen~\citep{codegen}, InCoder~\citep{incoder}, CodeLLaMa~\citep{codellama}, StarCoder~\citep{starcoder}, and DeepSeekCoder~\citep{deepseekcoder} using \ours. As~\Cref{tab:models} shows, these models vary in terms of parameters, data cutoff dates, open-source availability, and pretraining objectives. Given the multilingual (Python, Java, C++, and C\#) nature of \ours, our selection prioritizes models with multilingual capabilities, and exclude Python-only variants like CodeGen-Mono and StarCoder-Python. As we focus on code sources after April 2022, \ours guarantees that, with the exception of CodeLLaMa and DeepSeekCoder, all models are evaluated using clean, out-of-sample test cases. In \Cref{sec:appendix_data_contamination}, we further discuss the impact of data contamination on our evaluation results.

For GPT-3.5 and GPT-4, we use the OpenAI API for generation. For the remaining models, generation is conducted via the Huggingface \texttt{transformers} library, following established practices in \citet{incoder}, where we use top-p random sampling with \(p=0.95\) and a temperature of 0.2. Model details for reproducibility, including the model identifiers used on OpenAI API and the Huggingface model hub, are provided in \Cref{sec:model_names}.

%% file: tables/table_models.tex
\begin{table}[t]
\caption{Summary of evaluated models, highlighting data cutoff dates, open-source status (OS), and pretraining objectives. Dates in \textcolor{red}{red} indicate overlap between the model's pretraining data and the \ours benchmark in date range (post-April 2022). Data cutoff dates for InCoder are estimated based on their initial paper draft publication dates. The OS column denotes open-source availability ($\surd$ for yes, $\times$ for no), and the FIM column indicates models pretrained with FIM objectives and support for sentinel tokens in FIM inference. \textsuperscript{*}For CodeLLaMa, only 7B/13B versions support FIM inference, while the 34B version does not.}
\label{tab:models}
\vskip 0.15in
\begin{center}
\scalebox{0.823}{
\begin{tabular}{lllcc}
\toprule
               & \textbf{\#Params} & \textbf{Data Cutoff}            & \textbf{OS} & \textbf{FIM} \\ \midrule
GPT-3.5        & 175B              & Sept 2021                       & $\times$    & $\times$     \\
GPT-4          & -                 & Sept 2021                       & $\times$    & $\times$     \\
CodeGen        & 350M/2B/6B/16B    & Oct 2021                        & $\surd$     & $\times$     \\
InCoder        & 1.3B/6.7B         & $\le$ Mar 2022                  & $\surd$     & $\surd$      \\
CodeLLaMa      & 7B/13B/34B        & \textcolor{red}{Jul 2022} & $\surd$     & $\surd^*$      \\
StarCoder      & 15.5B             & Mar 2022                        & $\surd$     & $\surd$      \\
DeepSeekCoder  & 1.3B/6.7B/33B     & \textcolor{red}{Feb 2023}       & $\surd$     & $\surd$      \\ \bottomrule
\end{tabular}
}
\end{center}
\vskip -0.1in
\end{table}

%% file: contents/experimental_results.tex
\section{Experimental Results}\label{sec:experimental_results}

We now present the experimental results on our \ours benchmark, focusing on the effects of prompt designs, the efficacy of our syntax-aware truncation algorithm, and a comparative analysis of various LLMs across tasks. Given the inherent differences in model training environments and configurations, direct comparisons across different model families should be interpreted with caution. The primary value of our work is in establishing the \ours benchmark as a cornerstone for future experiments in this field.

\subsection{Impact of Prompt Designs}\label{sec:experimental_results_prompt}

\input{tables/table_prompts}

\Cref{tab:prompts} compares the effectiveness of different prompt designs by evaluating each model across various prompts with syntax-aware truncation in post-processing. This experiment reveals that:

\paragraph{Prompt Selection is Crucial for Fair Evaluation in Code FIM Tasks.} A narrow selection of prompt types can lead to skewed evaluation results, as different models respond differently due to differences in their pretraining data and methods. A potentially skewed evaluation by \citet{incoder} highlights this by comparing FIM models using the PSM prompt against non-FIM models with the IPF prompt. Doing so suggests a misleading superiority of InCoder-6B (25.2\%) over CodeGen-16B (15.2\%) in Pass@1 on \ours. This comparison, however, overlooks that CodeGen-16B achieves a higher Pass@1 of 25.9\% with the SPM prompt, a prompt not included in their evaluation setup. This example shows the necessity for a comprehensive prompt range to ensure fairness. Our work addresses this by reporting the best-performing prompt for each model and includes an extensive result table in \Cref{sec:all_results} for thorough comparison.

\paragraph{FIM Pretraining Boosts {\em Both} FIM and L2R Performance.} Pretraining LLMs with a FIM objective enhances their performance not only in FIM but also in left-to-right (L2R) generation. The advantage in FIM evaluation is highlighted by the results of CodeLLaMa models: the larger CodeLLaMa-34B, without FIM pretraining, is outperformed by the smaller, FIM+L2R pre-trained CodeLLaMa-13B. A more interesting observation emerges in the ``L2R'' column of \Cref{tab:prompts}: FIM-pretrained models like StarCoder outperform purely L2R-pretrained models like CodeGen-16B in L2R mode, despite similar sizes. This finding suggests that FIM pretraining does not harm, and actually enhances, a model's L2R performance, possibly by fostering a better understanding of code via contextually rich pretraining inputs. This supports similar improvements observed in FIM-pretrained GPT-3/Codex models in prior studies~\citep{humaneval_fim}, and offer strong justification for the recent shift from pure L2R pretraining to FIM pretraining among code LLM developers~\citep{starcoder,deepseekcoder,codellama}.

\subsection{Impact of Our Syntax-Aware Truncation}\label{sec:experimental_results_truncation}

\input{tables/table_truncation}

We assess the impact of our syntax-aware truncation algorithm through an ablation study, measuring model performance on the algorithmic block completion task with and without syntax-aware truncation. This analysis focuses on two key numbers: Pass@1 and the percentage of unexecutable programs due to compile or syntax errors in the generated completions. We treat empty outputs after truncation, typically indicative of a failure to identify any valid executable, as compilation errors. The results are shown in \Cref{tab:truncation}. These results show that:

\paragraph{Syntax-Aware Truncation Enhances FIM Output Quality.} \Cref{tab:truncation} shows that our syntax-aware truncation algorithm not only enhances the Pass@1 rates but also significantly reduces compilation errors across various models. This indicates a consistent improvement in the quality of FIM outputs, achieved without additional GPU overhead during model inference. We believe syntax-aware truncation holds promise for real-world code completion applications.

\paragraph{Syntax-Aware Truncation Enables Fair Comparison for Non-FIM Models.} As shown in \Cref{tab:truncation}, syntax-aware truncation benefits non-FIM models much more than FIM models. For example, CodeLLaMa-13B's Pass@1 rate jumps from 16.4\% to 41.4\% with truncation, changing its comparative performance against InCoder-6B, whose Pass@1 only increases marginally from 21.8\% to 25.2\%. This discrepancy stems from their distinct training approaches. InCoder, exclusively trained on FIM, naturally aligns with FIM-style prompts. In contrast, CodeLLaMa-13B, with a primary focus on L2R in its mixed FIM+L2R training, often produces unwanted extra code after completion. The extra code, while removable by syntax-aware truncation, obscures CodeLLaMa-13B's true effectiveness when such truncation is not applied. By precisely eliminating the extra code, syntax-aware truncation unveils the true coding proficiency of non-FIM or hybrid models like CodeLLaMa, ensuring fair comparisons with FIM-focused models. Additionally, syntax-aware truncation allows open-ended models to be evaluated in FIM tasks.

\subsection{Comparative Performance Analysis of LLMs}\label{sec:experimental_results_main}

\input{tables/table_main}

\input{figures/safim_average}

After determining the most effective prompt for each model and verifying the benefits of syntax-aware truncation, we conduct comprehensive evaluations across the entire \ours benchmark. \Cref{tab:main} shows model performances in each task category, and \Cref{fig:safim_average} visualizes the average performance of models against their model sizes. These results offers insights into the capabilities and limitations of code LLMs:

\paragraph{Pretraining Method and Data Are More Important Than Sheer Model Size.} Smaller models with sophisticated pretraining paradigms can match or even outperform larger counterparts. For example, StarCoder, with 15.5B parameters, achieves an average Pass@1 of 55.5\%, comparable to GPT-4's 53.3\%, despite GPT-4's vast size. This pattern recurs in models like CodeLLaMa-13B and DeepSeekCoder-1.3B. Notably, the comparison between StarCoder and GPT-4 is {\em not} subject to data contamination, as discussed in \Cref{tab:models}. This finding challenges the common belief that larger models automatically yield superior performance, even with basic pretraining methods~\citep{gpt3}. Our study suggests that this may not hold true for coding tasks: within the same model family, performance gains from increased size are only modest, while models from different families exhibit substantial performance variations. For example, the weakest CodeLLaMa model surpasses the strongest CodeGen model by 14 points, a far more significant margin than the 7.8-point spread within CodeLLaMa models.

\paragraph{Pretraining Method and Data Influence Task-Specific Performance.} We have discussed in \Cref{sec:experimental_results_prompt} that FIM pretraining enhances performance on both FIM evaluation and L2R completion. Dissecting model performance across \ours's three splits sheds further light on this impact:

\begin{itemize}[leftmargin=*,itemsep=2pt,topsep=2pt,parsep=2pt,partopsep=2pt]
\item For API function call completion, repository-level information is key. StarCoder and DeepSeekCoder, which excel in this task, both incorporate repository context into their pretraining data. StarCoder enriches its training input with GitHub issues and commit messages, while DeepSeekCoder organize code files according to their topological ordering based on API dependencies. These techniques significantly enhance their ability to understand API contexts.

\item For control-flow completion, CodeLLaMa's relatively strong performance is attributed to its use of execution-based feedback in its self-instruct training method. By executing generated code and applying the results as rewards or penalties, CodeLLaMa learns to avoid generating unexecutable code or infinite loops, thereby gaining a more refined understanding of control flows.
\end{itemize}

These findings highlight the pivotal role of the pretraining paradigm in the performance of LLMs on coding tasks.

%% file: tables/table_prompts.tex
\begin{table}[t]
\caption{Pass@1 of each model on algorithmic block completion, evaluated with various prompts and using syntax-aware truncation for post-processing. GPT-3.5, CodeGen-16B, and CodeLLaMa-34B cannot be evaluated with the Prefix-Suffix-Middle (PSM) prompt due to lack of support for FIM sentinel tokens, as discussed in \Cref{sec:prompts_and_post_processing_prompts}. The most effective prompt type for each model is highlighted in \textbf{bold}.}
\label{tab:prompts}
\vskip 0.15in
\begin{center}
\scalebox{0.84}{
\begin{tabular}{lccccc}
\toprule
& \textbf{L2R} & \textbf{PSM} & \textbf{SPM} & \textbf{IPF} & \textbf{1S} \\ \midrule

GPT-3.5 (175B) & \colorbox[HTML]{FFE0E0}{23.2} & - & \colorbox[HTML]{FFFDFD}{30.1} & \colorbox[HTML]{FFF6F6}{28.6} & \colorbox[HTML]{FCFFFC}{\textbf{31.2}} \\
CodeGen-16B & \colorbox[HTML]{FFE6E6}{24.6} & - & \colorbox[HTML]{FFEBEB}{\textbf{25.9}} & \colorbox[HTML]{FFBEBE}{15.2} & \colorbox[HTML]{FF8080}{\hspace{5pt}0.4} \\
InCoder-6B & \colorbox[HTML]{FFCACA}{18.1} & \colorbox[HTML]{FFE8E8}{\textbf{25.2}} & \colorbox[HTML]{FFE3E3}{24.1} & \colorbox[HTML]{FFB1B1}{12.2} & \colorbox[HTML]{FFE0E0}{23.2} \\
CodeLLaMa-13B & \colorbox[HTML]{F8FFF8}{32.3} & \colorbox[HTML]{FFA9A9}{10.2} & \colorbox[HTML]{D1FFD1}{\textbf{41.4}} & \colorbox[HTML]{FEFFFE}{30.9} & \colorbox[HTML]{FFC2C2}{16.1} \\
CodeLLaMa-34B & \colorbox[HTML]{EAFFEA}{35.5} & - & \colorbox[HTML]{DDFFDD}{\textbf{38.5}} & \colorbox[HTML]{EBFFEB}{35.4} & \colorbox[HTML]{FFD1D1}{19.6} \\
StarCoder (15.5B) & \colorbox[HTML]{FFFAFA}{29.3} & \colorbox[HTML]{C6FFC6}{44.0} & \colorbox[HTML]{C6FFC6}{\textbf{44.1}} & \colorbox[HTML]{FFD6D6}{20.8} & \colorbox[HTML]{CDFFCD}{42.4} \\
DeepSeekCoder-33B & \colorbox[HTML]{D0FFD0}{41.6} & \colorbox[HTML]{80FF80}{\textbf{60.8}} & \colorbox[HTML]{8EFF8E}{57.4} & \colorbox[HTML]{F1FFF1}{33.8} & \colorbox[HTML]{83FF83}{59.9} \\

\bottomrule

\end{tabular}
}
\end{center}
\vskip -0.1in
\end{table}

%% file: tables/table_truncation.tex
\begin{table}[t]
\caption{Comparison of model performance with and without our syntax-aware truncation algorithm in the post-processing phase. This table presents two numbers for each model evaluated on algorithmic block completion tasks: \textbf{Pass@1} and \textbf{CErr\%} (the percentage of unexecutable programs due to compile or syntax errors in the generated completions).}
\label{tab:truncation}
\vskip 0.15in
\begin{center}
\scalebox{0.83}{
\begin{tabular}{lcccc}
\toprule
& \multicolumn{2}{c}{\textbf{No Trunc.}} & \multicolumn{2}{c}{\textbf{Syntax Trunc.}} \\
\cmidrule(lr){2-3} \cmidrule(lr){4-5}
& \textbf{Pass@1} & \textbf{CErr\%} & \textbf{Pass@1} & \textbf{CErr\%} \\ \midrule

GPT-3.5 (175B) & \colorbox[HTML]{FFF8F8}{28.7} & \colorbox[HTML]{FFE3E3}{25.3} & \colorbox[HTML]{FBFFFB}{31.2} & \colorbox[HTML]{FFEEEE}{17.0} \\
GPT-4 ($>$ 220B) & \colorbox[HTML]{D0FFD0}{41.7} & \colorbox[HTML]{FFE3E3}{25.4} & \colorbox[HTML]{CEFFCE}{42.1} & \colorbox[HTML]{FFE6E6}{22.9} \\
CodeGen-16B & \colorbox[HTML]{FF8080}{\hspace{5pt}0.0} & \colorbox[HTML]{FF8080}{99.9} & \colorbox[HTML]{FFECEC}{25.9} & \colorbox[HTML]{FFEDED}{17.9} \\
InCoder-6B & \colorbox[HTML]{FFDBDB}{21.8} & \colorbox[HTML]{FFE2E2}{25.7} & \colorbox[HTML]{FFE9E9}{25.2} & \colorbox[HTML]{FFF3F3}{13.2} \\
CodeLLaMa-13B & \colorbox[HTML]{FFC4C4}{16.4} & \colorbox[HTML]{FFAEAE}{64.6} & \colorbox[HTML]{D1FFD1}{41.4} & \colorbox[HTML]{FFF6F6}{10.9} \\
CodeLLaMa-34B & \colorbox[HTML]{FF8484}{\hspace{5pt}1.0} & \colorbox[HTML]{FF8787}{94.5} & \colorbox[HTML]{DDFFDD}{38.5} & \colorbox[HTML]{FFF1F1}{14.7} \\
StarCoder (15.5B) & \colorbox[HTML]{CBFFCB}{42.7} & \colorbox[HTML]{FFF1F1}{14.3} & \colorbox[HTML]{C5FFC5}{44.1} & \colorbox[HTML]{FFF8F8}{\hspace{5pt}9.5} \\
DeepSeekCoder-33B & \colorbox[HTML]{84FF84}{59.7} & \colorbox[HTML]{FFFAFA}{\hspace{5pt}8.0} & \colorbox[HTML]{80FF80}{60.8} & \colorbox[HTML]{FFFFFF}{\hspace{5pt}4.0} \\

\bottomrule
\end{tabular}
}
\end{center}
\vskip -0.1in
\end{table}

%% file: tables/table_main.tex
\begin{table}[t]
\caption{Pass@1 of various models on the \ours benchmark, showing their performance in algorithmic block completion (Algo.), control-flow completion (Control), and API function call completion (API). The table also reports the average performance, indicating each model's overall effectiveness on \ours.}
\label{tab:main}
\vskip 0.15in
\begin{center}
\scalebox{0.9}{
\begin{tabular}{lcccc}
\toprule
& \textbf{Algo.} & \textbf{Control} & \textbf{API} & \textbf{Avg}
\\ \midrule

GPT-3.5 (175B) & \colorbox[HTML]{FFD5D5}{31.2} & \colorbox[HTML]{FFCDCD}{37.5} & \colorbox[HTML]{EFFFEF}{53.9} & \colorbox[HTML]{FFE3E3}{40.9} \\
GPT-4 ($>$ 220B) & \colorbox[HTML]{EBFFEB}{42.1} & \colorbox[HTML]{D4FFD4}{55.2} & \colorbox[HTML]{C1FFC1}{62.6} & \colorbox[HTML]{D7FFD7}{53.3} \\
CodeGen-350M & \colorbox[HTML]{FF8080}{16.3} & \colorbox[HTML]{FF9090}{26.1} & \colorbox[HTML]{FF8080}{26.5} & \colorbox[HTML]{FF8080}{22.9} \\
CodeGen-2B & \colorbox[HTML]{FFA9A9}{23.5} & \colorbox[HTML]{FFB4B4}{32.9} & \colorbox[HTML]{FF9E9E}{32.3} & \colorbox[HTML]{FFA4A4}{29.5} \\
CodeGen-6B & \colorbox[HTML]{FFA9A9}{23.6} & \colorbox[HTML]{FFBFBF}{34.8} & \colorbox[HTML]{FF8686}{27.7} & \colorbox[HTML]{FF9F9F}{28.7} \\
CodeGen-16B & \colorbox[HTML]{FFB7B7}{25.9} & \colorbox[HTML]{FFC3C3}{35.7} & \colorbox[HTML]{FF9999}{31.3} & \colorbox[HTML]{FFACAC}{31.0} \\
InCoder-1B & \colorbox[HTML]{FF9B9B}{21.1} & \colorbox[HTML]{FF8080}{22.9} & \colorbox[HTML]{FFDBDB}{43.9} & \colorbox[HTML]{FFA3A3}{29.3} \\
InCoder-6B & \colorbox[HTML]{FFB2B2}{25.2} & \colorbox[HTML]{FF9B9B}{28.2} & \colorbox[HTML]{FFF1F1}{48.1} & \colorbox[HTML]{FFBCBC}{33.8} \\
CodeLLaMa-7B & \colorbox[HTML]{FFE9E9}{34.7} & \colorbox[HTML]{DCFFDC}{53.6} & \colorbox[HTML]{FFEAEA}{46.8} & \colorbox[HTML]{FFFAFA}{45.0} \\
CodeLLaMa-13B & \colorbox[HTML]{EFFFEF}{41.4} & \colorbox[HTML]{C9FFC9}{57.2} & \colorbox[HTML]{D1FFD1}{59.7} & \colorbox[HTML]{D9FFD9}{52.8} \\
CodeLLaMa-34B & \colorbox[HTML]{FFFFFF}{38.5} & \colorbox[HTML]{DAFFDA}{54.0} & \colorbox[HTML]{E1FFE1}{56.5} & \colorbox[HTML]{EBFFEB}{49.7} \\
StarCoder (15.5B) & \colorbox[HTML]{DFFFDF}{44.1} & \colorbox[HTML]{D8FFD8}{54.5} & \colorbox[HTML]{A5FFA5}{68.1} & \colorbox[HTML]{CAFFCA}{55.5} \\
DeepSeekCoder-1.3B & \colorbox[HTML]{F0FFF0}{41.2} & \colorbox[HTML]{D9FFD9}{54.1} & \colorbox[HTML]{C1FFC1}{62.6} & \colorbox[HTML]{DAFFDA}{52.6} \\
DeepSeekCoder-6.7B & \colorbox[HTML]{A2FFA2}{54.7} & \colorbox[HTML]{9CFF9C}{65.8} & \colorbox[HTML]{9CFF9C}{69.7} & \colorbox[HTML]{9FFF9F}{63.4} \\
DeepSeekCoder-33B & \colorbox[HTML]{80FF80}{\textbf{60.8}} & \colorbox[HTML]{80FF80}{\textbf{71.1}} & \colorbox[HTML]{80FF80}{\textbf{75.2}} & \colorbox[HTML]{80FF80}{\textbf{69.0}} \\

\bottomrule
\end{tabular}
}
\end{center}
\vskip -0.1in
\end{table}

%% file: figures/safim_average.tex
\begin{figure}[t]
\vskip 0.2in
\begin{center}
\centerline{\includegraphics{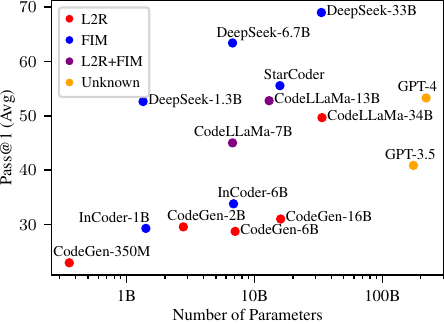}}
\caption{Average performance of different models relative to their sizes on the \ours benchmark. Each model is represented by a dot, with the x-axis showing model size (number of parameters) and the y-axis showing average performance across three task categories. Dot colors signify pretraining paradigms: \textcolor{red}{red} for Left-to-Right (L2R), \textcolor{blue}{blue} for FIM, \textcolor{purple}{purple} for a combination of L2R and FIM, and \textcolor{orange}{orange} for proprietary models with undisclosed pretraining methods.}
\label{fig:safim_average}
\end{center}
\vskip -0.2in
\end{figure}

%% file: contents/conclusion.tex
\section{Conclusion and Future Work}\label{sec:conclusion_and_future_work}

We introduced the Syntax-Aware Fill-in-the-Middle (\ours) benchmark, the first large-scale, multilingual Fill-in-the-Middle (FIM) benchmark equipped with executable unit tests for evaluating code-centric Large Language Models (LLMs). To mitigate data contamination, \ours adopts a strict cutoff date for code sources. Moreover, \ours uniquely categorizes tasks into three syntax-driven splits: algorithmic block completion, control-flow expression completion, and API function call completion. These splits provide a comprehensive assessment of LLMs' coding capabilities across multiple dimensions. \ours's suite of prompts and its novel syntax-aware truncation algorithm for post-processing enable fair comparisons among various types of models, including those not explicitly pretrained on FIM tasks.

The results of our large-scale evaluation highlight the significant impact of pretraining paradigms on LLMs' performance, emphasizing the importance of training method and data quality over sheer model size. 
We found that FIM pretraining can enhance, rather than harm, Left-to-Right (L2R) inference capabilities, supporting a shift towards FIM as a primary pretraining objective for code LLMs. We acknowledge a key limitation in our study: our conclusions are drawn from comparisons across various model families trained with different paradigms, rather than from controlled experiments altering pretraining paradigms within the same model. Yet, \ours establishes a foundational framework for future research into pretraining paradigms and the development of better LLMs for coding tasks.

%% file: contents/acknowledgements.tex
\section*{Acknowledgements}

We thank the reviewers at ICML for their constructive feedback and support. This work is supported in part by gift from Meta, the U.S. National Science Foundation through grants IIS-1955488, IIS-2027575, ARO W911NF2110339, ONR N00014-21-1-2724, and DOE awards DE-SC0016260, DE-SC0021982. We thank Chenyan Xiong and Carlton Shen for providing computational resources that enabled us to conduct urgent experiments during the rebuttal period.

%% file: contents/broader_impact.tex
\section*{Impact Statement}

In this paper, we introduce Syntax-Aware Fill-in-the-Middle (\ours), a benchmark aimed at enhancing the capabilities of Large Language Models (LLMs) in code generation tasks. The advancement of LLMs in code generation raises concerns about automated code production's security, privacy, and potential misuse. There is a risk that improved code generation capabilities could be exploited for malicious purposes, such as automating the creation of software vulnerabilities or facilitating the development of harmful software. Our research emphasizes the importance of responsible AI development and use, advocating for continuous monitoring, ethical guidelines, and safeguards to mitigate these risks.

%% file: contents/appendix/benchmark_api_libraries.tex
\subsection{Details about the API Function Call Completion Task}\label{sec:benchmark_api_libraries}

We consider the following API libraries for each programming language when we construct the API function call completion split of \ours:

\begin{itemize}
\item \textbf{Python:} NumPy, Pandas, Statsmodels, Sci-kit Learn, Matplotlib, NLTK, Gensim, XGBoost, PyTorch, Huggingface Transformers
\item \textbf{Java:} GSON, Caffeine, Apache Commons, Google HTTP Client, Joda-Time, JavaParser, 
\item \textbf{C++:} GMP, Boost, JSON, QT, Eigen, OpenGL, Tree-Sitter
\item \textbf{C\#:} Newtonsoft.Json, SignalR, RestSharp, LiteDB, BCrypt.Net
\end{itemize}

%% file: contents/appendix/benchmark_statistics.tex
\subsection{Statistics of the \ours Benchmark}\label{sec:benchmark_statistics}

Statistics of each split of the \ours benchmark is presented in \Cref{tab:appendix_statistics}.

\input{tables/table_appendix_statistics}

\input{figures/data_lengths}

\Cref{fig:data_lengths} shows the distribution the total lengths of problem descriptions plus code contexts of examples measured in characters. A majority of the dataset has less than 6k characters. On average, one BPE token corresponds to 3 to 4 characters in the code domain. This ensures that the evaluated models, all with context windows of at least 2,048, accurately reflect performance without bias from input size.

\input{tables/table_appendix_statistics_per_pl}

\Cref{tab:appendix_statistics_per_pl} shows statistics per programming language of examples in \ours. This distribution of PLs in \ours mirrors the prevalence in our source corpus, especially in Codeforces where most contestants use C++. \Cref{tab:appendix_statistics_per_pl} also highlights variations in coding style across languages. For instance, C++ and Python programmers favor succinct coding (less code, shorter variable names), while Java and C\# users lean towards verbosity. Subsequent sections will discuss how these differences in coding style influence evaluation results.

%% file: tables/table_appendix_statistics.tex
\begin{table}[h]
\caption{Statistics of each task category of the \ours benchmark, including number of examples, total uncompressed disk size of code contexts, average length of code contexts in bytes, and average length of ground truth completions in bytes.}
\label{tab:appendix_statistics}
\vskip 0.15in
\begin{center}
\scalebox{0.9}{
\begin{tabular}{lrrrrr}
\toprule
 & \textbf{Source} & \textbf{\# Examples} & \textbf{Disk Size} & \textbf{Avg Code Len} & \textbf{Avg Completion Len} \\ \midrule
Algorithmic Block & Codeforces & 8,781 & 29.4M & 3346B & 67B \\
Control-Flow & Codeforces & 8,629 & 29.5M & 3415B & 16B \\ 
API Function Call & GitHub & 310 & 713K & 2302B & 40B \\ \midrule
Total & - & 17,720 & 59.6M & 3364B & 42B \\
\bottomrule
\end{tabular}
}
\end{center}
\vskip -0.1in
\end{table}

%% file: figures/data_lengths.tex
\begin{figure}[h]
\vskip 0.2in
\begin{center}
\centerline{\includegraphics{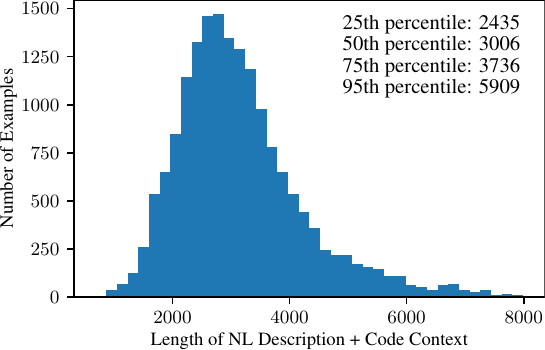}}
\caption{Histogram of the total number of \textbf{characters} of the natural language problem description and the code context. 424 example longer than 8,000 characters are excluded from this histogram for clarity but counted towards the displayed quantiles.}
\label{fig:data_lengths}
\end{center}
\vskip -0.2in
\end{figure}

%% file: tables/table_appendix_statistics_per_pl.tex
\begin{table}[t]
\caption{Statistics of examples in each programming language of the \ours benchmark, including number of examples, total uncompressed disk size of code contexts, average length of code contexts in bytes,  average length of ground truth completions in bytes, and average length of identifiers in bytes. The \textit{identifiers} refer to the names of variables, functions, and classes.}
\label{tab:appendix_statistics_per_pl}
\vskip 0.15in
\begin{center}
\scalebox{0.9}{
\begin{tabular}{lrrrrr}
\toprule
 & \textbf{\# Examples} & \textbf{Disk Size} & \textbf{Avg Code Len} & \textbf{Avg Completion Len} & \textbf{Avg Identifier Len} \\ \midrule
Python & 9,901 & 30.0M & 3026B & 44B & 2.73B \\
Java & 4,999 & 17.3M & 3454B & 44B & 4.14B \\
C++ & 1,736 & 5.14M & 2962B & 27B & 3.65B \\
C\# & 1,084 & 7.24M & 6675B & 42B & 5.79B \\ \midrule
Total & 17,720 & 59.6M & 3364B & 42B & 3.59B \\
\bottomrule
\end{tabular}
}
\end{center}
\vskip -0.1in
\end{table}

%% file: contents/appendix/model_names.tex
\subsection{Details of Model Implementations}\label{sec:model_names}

\Cref{tab:models} shows the implementations used for evaluating each LLM, including additional models we will discuss in \Cref{sec:additional_main_results}. For GPT-3.5 and GPT-4, we use the OpenAI API\footnote{\url{https://openai.com/blog/openai-api}} for generation. For the remaining models, generation is conducted via the Huggingface \texttt{transformers} library\footnote{\url{https://github.com/huggingface/transformers}}.

\input{tables/table_appendix_models}

%% file: tables/table_appendix_models.tex
\begin{table}[h]
\caption{The code enviroment for evaluating each LLM and the model identifier on its respective platform.}
\label{tab:appendix_models}
\vskip 0.15in
\begin{center}
\scalebox{0.9}{
\begin{tabular}{lll}
\toprule
& \textbf{Codebase} & \textbf{Model Identifier} \\ \midrule
GPT-3.5 & OpenAI API & \texttt{gpt-3.5-turbo-0301} \\
GPT-4 & OpenAI API & \texttt{gpt-4-1106-preview} \\
CodeGen-350M & Huggingface Transformers & \texttt{Salesforce/codegen-350M-multi} \\
CodeGen-2B & Huggingface Transformers & \texttt{Salesforce/codegen-2B-multi} \\
CodeGen-6B & Huggingface Transformers & \texttt{Salesforce/codegen-6B-multi} \\
CodeGen-16B & Huggingface Transformers & \texttt{Salesforce/codegen-16B-multi} \\
InCoder-1B & Huggingface Transformers & \texttt{facebook/incoder-1B} \\
InCoder-6B & Huggingface Transformers & \texttt{facebook/incoder-6B} \\
CodeLLaMa-7B & Huggingface Transformers & \texttt{codellama/CodeLlama-7b-hf} \\
CodeLLaMa-13B & Huggingface Transformers & \texttt{codellama/CodeLlama-13b-hf} \\
CodeLLaMa-34B & Huggingface Transformers & \texttt{codellama/CodeLlama-34b-hf} \\
StarCoder (15.5B) & Huggingface Transformers & \texttt{bigcode/starcoderbase} \\
DeepSeekCoder-1.3B & Huggingface Transformers & \texttt{deepseek-ai/deepseek-coder-1.3b-base} \\
DeepSeekCoder-6.7B & Huggingface Transformers & \texttt{deepseek-ai/deepseek-coder-6.7b-base} \\
DeepSeekCoder-33B & Huggingface Transformers & \texttt{deepseek-ai/deepseek-coder-33b-base} \\
Mixtral-8x7B & Huggingface Transformers & \texttt{mistralai/Mixtral-8x7B-v0.1} \\
Phi-1.5 (1.3B) & Huggingface Transformers & \texttt{microsoft/phi-1\_5} \\
Phi-2 (2.7B) & Huggingface Transformers & \texttt{microsoft/phi-2} \\
WizardCoder-1B & Huggingface Transformers & \texttt{WizardLM/WizardCoder-1B-V1.0} \\
WizardCoder-3B & Huggingface Transformers & \texttt{WizardLM/WizardCoder-3B-V1.0} \\
WizardCoder-15B & Huggingface Transformers & \texttt{WizardLM/WizardCoder-15B-V1.0} \\
WizardCoder-33B & Huggingface Transformers & \texttt{WizardLM/WizardCoder-33B-V1.1} \\
Magicoder-S-DS-6.7B & Huggingface Transformers & \texttt{ise-uiuc/Magicoder-S-DS-6.7B} \\
\bottomrule
\end{tabular}
}
\end{center}
\vskip -0.1in
\end{table}

%% file: contents/appendix/all_results.tex
\subsection{Results of All Models on All Prompts}\label{sec:all_results}

\Cref{tab:appendix_extra_results_block}, \Cref{tab:appendix_extra_results_control}, and \Cref{tab:appendix_extra_results_api} show experimental results of all models using all types of prompts, where each table shows the results on one task category of \ours.

\input{tables/table_appendix_extra_results_block}

\input{tables/table_appendix_extra_results_control}

\input{tables/table_appendix_extra_results_api}

%% file: tables/table_appendix_extra_results_block.tex
\begin{table}[htb!]
\caption{The performance of each model with each type of prompts on algorithmic block completion. Syntax-aware truncation is used for post-processing. The most effective prompt type for each model is highlighted in \textbf{bold}.}
\label{tab:appendix_extra_results_block}
\vskip 0.15in
\begin{center}
\scalebox{0.9}{
\begin{tabular}{lcccccc}
\toprule
& \textbf{L2R} & \textbf{PSM} & \textbf{SPM} & \textbf{IPF} & \textbf{1S} \\ \midrule

GPT-3.5 (175B) & \colorbox[HTML]{FFE1E1}{23.2} & - & \colorbox[HTML]{FFFEFE}{30.1} & \colorbox[HTML]{FFF7F7}{28.6} & \colorbox[HTML]{FBFFFB}{\textbf{31.2}} \\
GPT-4 & - & - & - & - & \colorbox[HTML]{CEFFCE}{\textbf{42.1}} \\
CodeGen-350M & \colorbox[HTML]{FFC0C0}{15.4} & - & \colorbox[HTML]{FFC4C4}{\textbf{16.3}} & \colorbox[HTML]{FF9C9C}{\hspace{5pt}6.8} & \colorbox[HTML]{FF8080}{\hspace{5pt}0.1} \\
CodeGen-2B & \colorbox[HTML]{FFDEDE}{22.5} & - & \colorbox[HTML]{FFE2E2}{\textbf{23.5}} & \colorbox[HTML]{FFBABA}{13.9} & \colorbox[HTML]{FF8080}{\hspace{5pt}0.0} \\
CodeGen-6B & \colorbox[HTML]{FFE1E1}{23.2} & - & \colorbox[HTML]{FFE2E2}{\textbf{23.6}} & \colorbox[HTML]{FFBDBD}{14.6} & \colorbox[HTML]{FF8080}{\hspace{5pt}0.0} \\
CodeGen-16B & \colorbox[HTML]{FFE7E7}{24.6} & - & \colorbox[HTML]{FFECEC}{\textbf{25.9}} & \colorbox[HTML]{FFBFBF}{15.2} & \colorbox[HTML]{FF8181}{\hspace{5pt}0.4} \\
InCoder-1B & \colorbox[HTML]{FFBBBB}{14.1} & \colorbox[HTML]{FFD8D8}{\textbf{21.1}} & \colorbox[HTML]{FFD0D0}{19.2} & \colorbox[HTML]{FFA5A5}{\hspace{5pt}9.0} & \colorbox[HTML]{FFC9C9}{17.6} \\
InCoder-6B & \colorbox[HTML]{FFCBCB}{18.1} & \colorbox[HTML]{FFE9E9}{\textbf{25.2}} & \colorbox[HTML]{FFE4E4}{24.1} & \colorbox[HTML]{FFB3B3}{12.2} & \colorbox[HTML]{FFE1E1}{23.2} \\
CodeLLaMa-7B & \colorbox[HTML]{FEFFFE}{30.7} & \colorbox[HTML]{FFA5A5}{\hspace{5pt}8.8} & \colorbox[HTML]{EDFFED}{\textbf{34.7}} & \colorbox[HTML]{FFE6E6}{24.4} & \colorbox[HTML]{FF9F9F}{\hspace{5pt}7.5} \\
CodeLLaMa-13B & \colorbox[HTML]{F7FFF7}{32.3} & \colorbox[HTML]{FFAAAA}{10.2} & \colorbox[HTML]{D1FFD1}{\textbf{41.4}} & \colorbox[HTML]{FDFFFD}{30.9} & \colorbox[HTML]{FFC3C3}{16.1} \\
CodeLLaMa-34B & \colorbox[HTML]{EAFFEA}{35.5} & - & \colorbox[HTML]{DDFFDD}{\textbf{38.5}} & \colorbox[HTML]{EAFFEA}{35.4} & \colorbox[HTML]{FFD2D2}{19.6} \\
StarCoder (15.5B) & \colorbox[HTML]{FFFAFA}{29.3} & \colorbox[HTML]{C6FFC6}{44.0} & \colorbox[HTML]{C5FFC5}{\textbf{44.1}} & \colorbox[HTML]{FFD7D7}{20.8} & \colorbox[HTML]{CDFFCD}{42.4} \\
DeepSeekCoder-1.3B & \colorbox[HTML]{FFF5F5}{28.0} & \colorbox[HTML]{D2FFD2}{\textbf{41.2}} & \colorbox[HTML]{DCFFDC}{38.7} & \colorbox[HTML]{FF9B9B}{\hspace{5pt}6.5} & \colorbox[HTML]{DFFFDF}{38.0} \\
DeepSeekCoder-6.7B & \colorbox[HTML]{E7FFE7}{36.2} & \colorbox[HTML]{99FF99}{\textbf{54.7}} & \colorbox[HTML]{A7FFA7}{51.3} & \colorbox[HTML]{FFF1F1}{27.1} & \colorbox[HTML]{A0FFA0}{52.9} \\
DeepSeekCoder-33B & \colorbox[HTML]{D0FFD0}{41.6} & \colorbox[HTML]{80FF80}{\textbf{60.8}} & \colorbox[HTML]{8EFF8E}{57.4} & \colorbox[HTML]{F1FFF1}{33.8} & \colorbox[HTML]{83FF83}{59.9} \\
\bottomrule

\end{tabular}
}
\end{center}
\vskip -0.1in
\end{table}

%% file: tables/table_appendix_extra_results_control.tex
\begin{table}[htb!]
\caption{The performance of each model with each type of prompts on control-flow completion. Syntax-aware truncation is used for post-processing. The most effective prompt type for each model is highlighted in \textbf{bold}.}
\label{tab:appendix_extra_results_control}
\vskip 0.15in
\begin{center}
\scalebox{0.9}{
\begin{tabular}{lcccccc}
\toprule
& \textbf{L2R} & \textbf{PSM} & \textbf{SPM} & \textbf{IPF} & \textbf{1S} \\ \midrule

GPT-3.5 (175B) & - & - & - & - & \colorbox[HTML]{FFEFEF}{\textbf{37.5}} \\
GPT-4 & - & - & - & - & \colorbox[HTML]{C4FFC4}{\textbf{55.2}} \\
CodeGen-350M & \colorbox[HTML]{FFB9B9}{25.0} & - & \colorbox[HTML]{FFBEBE}{\textbf{26.1}} & \colorbox[HTML]{FF9A9A}{17.6} & - \\
CodeGen-2B & \colorbox[HTML]{FFD9D9}{32.4} & - & \colorbox[HTML]{FFDBDB}{\textbf{32.9}} & \colorbox[HTML]{FFBABA}{25.1} & - \\
CodeGen-6B & \colorbox[HTML]{FFDCDC}{33.1} & - & \colorbox[HTML]{FFE3E3}{\textbf{34.8}} & \colorbox[HTML]{FFBDBD}{25.9} & - \\
CodeGen-16B & \colorbox[HTML]{FFE3E3}{34.7} & - & \colorbox[HTML]{FFE7E7}{\textbf{35.7}} & \colorbox[HTML]{FFC6C6}{27.9} & - \\
InCoder-1B & \colorbox[HTML]{FFA2A2}{19.6} & \colorbox[HTML]{FFB0B0}{22.9} & \colorbox[HTML]{FFB7B7}{\textbf{24.4}} & \colorbox[HTML]{FF8080}{11.5} & - \\
InCoder-6B & \colorbox[HTML]{FFB3B3}{23.6} & \colorbox[HTML]{FFC7C7}{28.2} & \colorbox[HTML]{FFCACA}{\textbf{29.0}} & \colorbox[HTML]{FF8E8E}{14.9} & - \\
CodeLLaMa-7B & \colorbox[HTML]{F7FFF7}{43.1} & \colorbox[HTML]{FFBDBD}{25.8} & \colorbox[HTML]{CAFFCA}{\textbf{53.6}} & \colorbox[HTML]{FFFCFC}{40.6} & - \\
CodeLLaMa-13B & \colorbox[HTML]{EFFFEF}{45.1} & \colorbox[HTML]{FFC3C3}{27.3} & \colorbox[HTML]{BBFFBB}{\textbf{57.2}} & \colorbox[HTML]{EAFFEA}{46.2} & - \\
CodeLLaMa-34B & \colorbox[HTML]{E2FFE2}{48.0} & - & \colorbox[HTML]{C9FFC9}{\textbf{54.0}} & \colorbox[HTML]{D3FFD3}{51.5} & - \\
StarCoder (15.5B) & \colorbox[HTML]{F6FFF6}{43.4} & \colorbox[HTML]{C7FFC7}{\textbf{54.5}} & \colorbox[HTML]{CAFFCA}{53.7} & \colorbox[HTML]{FFEEEE}{37.4} & - \\
DeepSeekCoder-1.3B & \colorbox[HTML]{F9FFF9}{42.6} & \colorbox[HTML]{C8FFC8}{\textbf{54.1}} & \colorbox[HTML]{CFFFCF}{52.5} & \colorbox[HTML]{FFE5E5}{35.1} & - \\
DeepSeekCoder-6.7B & \colorbox[HTML]{D8FFD8}{50.4} & \colorbox[HTML]{96FF96}{\textbf{65.8}} & \colorbox[HTML]{9FFF9F}{63.8} & \colorbox[HTML]{D4FFD4}{51.4} & - \\
DeepSeekCoder-33B & \colorbox[HTML]{C1FFC1}{55.7} & \colorbox[HTML]{80FF80}{\textbf{71.1}} & \colorbox[HTML]{85FF85}{69.8} & \colorbox[HTML]{B5FFB5}{58.6} & - \\

\bottomrule

\end{tabular}
}
\end{center}
\vskip -0.1in
\end{table}

%% file: tables/table_appendix_extra_results_api.tex
\begin{table}[htb!]
\caption{The performance of each model with each type of prompts on API function call completion. Syntax-aware truncation is used for post-processing. The most effective prompt type for each model is highlighted in \textbf{bold}.}
\label{tab:appendix_extra_results_api}
\vskip 0.15in
\begin{center}
\scalebox{0.9}{
\begin{tabular}{lcccccc}
\toprule
& \textbf{L2R} & \textbf{PSM} & \textbf{SPM} & \textbf{IPF} & \textbf{1S} \\ \midrule

GPT-3.5 (175B) & - & - & - & - & \colorbox[HTML]{D2FFD2}{\textbf{53.9}} \\
GPT-4 & - & - & - & - & \colorbox[HTML]{B0FFB0}{\textbf{62.6}} \\
CodeGen-350M & \colorbox[HTML]{FFB6B6}{23.5} & - & \colorbox[HTML]{FFC1C1}{\textbf{26.5}} & \colorbox[HTML]{FF8080}{\hspace{5pt}9.7} & - \\
CodeGen-2B & \colorbox[HTML]{FFD0D0}{30.3} & - & \colorbox[HTML]{FFD7D7}{\textbf{32.3}} & \colorbox[HTML]{FF8282}{10.3} & - \\
CodeGen-6B & \colorbox[HTML]{FFBDBD}{25.5} & - & \colorbox[HTML]{FFC6C6}{\textbf{27.7}} & \colorbox[HTML]{FF8F8F}{13.5} & - \\
CodeGen-16B & \colorbox[HTML]{FFD4D4}{\textbf{31.3}} & - & \colorbox[HTML]{FFD4D4}{31.3} & \colorbox[HTML]{FF9B9B}{16.8} & - \\
InCoder-1B & \colorbox[HTML]{FFEFEF}{38.4} & \colorbox[HTML]{F9FFF9}{\textbf{43.9}} & \colorbox[HTML]{F9FFF9}{43.9} & \colorbox[HTML]{FF8F8F}{13.5} & - \\
InCoder-6B & \colorbox[HTML]{FFF9F9}{41.0} & \colorbox[HTML]{E9FFE9}{\textbf{48.1}} & \colorbox[HTML]{EDFFED}{47.1} & \colorbox[HTML]{FF9A9A}{16.5} & - \\
CodeLLaMa-7B & \colorbox[HTML]{E7FFE7}{\textbf{48.7}} & \colorbox[HTML]{FFEAEA}{37.1} & \colorbox[HTML]{EEFFEE}{46.8} & \colorbox[HTML]{FFAEAE}{21.6} & - \\
CodeLLaMa-13B & \colorbox[HTML]{E0FFE0}{50.3} & \colorbox[HTML]{FFF2F2}{39.0} & \colorbox[HTML]{BCFFBC}{\textbf{59.7}} & \colorbox[HTML]{FFF2F2}{39.0} & - \\
CodeLLaMa-34B & \colorbox[HTML]{DFFFDF}{50.6} & - & \colorbox[HTML]{EAFFEA}{47.7} & \colorbox[HTML]{C8FFC8}{\textbf{56.5}} & - \\
StarCoder (15.5B) & \colorbox[HTML]{DFFFDF}{50.6} & \colorbox[HTML]{9BFF9B}{\textbf{68.1}} & \colorbox[HTML]{A6FFA6}{65.2} & \colorbox[HTML]{F7FFF7}{44.5} & - \\
DeepSeekCoder-1.3B & \colorbox[HTML]{F2FFF2}{45.8} & \colorbox[HTML]{B0FFB0}{\textbf{62.6}} & \colorbox[HTML]{DAFFDA}{51.9} & \colorbox[HTML]{FF8888}{11.9} & - \\
DeepSeekCoder-6.7B & \colorbox[HTML]{D9FFD9}{52.3} & \colorbox[HTML]{95FF95}{\textbf{69.7}} & \colorbox[HTML]{BBFFBB}{60.0} & \colorbox[HTML]{D9FFD9}{52.3} & - \\
DeepSeekCoder-33B & \colorbox[HTML]{F3FFF3}{45.5} & \colorbox[HTML]{80FF80}{\textbf{75.2}} & \colorbox[HTML]{A9FFA9}{64.5} & \colorbox[HTML]{DFFFDF}{50.6} & - \\

\bottomrule

\end{tabular}
}
\end{center}
\vskip -0.1in
\end{table}

%% file: contents/appendix/extra_results_truncation.tex
\subsection{Further Results about Syntax-Aware Truncation}\label{sec:extra_results_truncation}

\Cref{sec:experimental_results_truncation} explores the benefits of syntax-aware truncation with algorithmic block completion tasks. This section extends the results in \Cref{tab:truncation} to encompass all tasks in \ours. Additionally, we also show the selected prompt for each model, determined by the highest pass@1 rate post-truncation. The results are shown in \Cref{tab:appendix_truncation_block}, \Cref{tab:appendix_truncation_control}, and \Cref{tab:appendix_truncation_api}.

\input{tables/table_appendix_trunc_results_block}

\input{tables/table_appendix_trunc_results_control}

\input{tables/table_appendix_trunc_results_api}

We find that syntax-aware truncation consistently improves the pass@1 rate and reduces compilation errors in both algorithmic block completion and control-flow expression completion, highlighting the effectiveness of syntax-aware truncation.

However, in API function call completion, which involves generation of function invocation expressions or statements, LLMs typically produce error-free code without truncation, as these code segments are typically short and naturally segmented with line breaks. That said, syntax-aware truncation becomes crucial for models and prompts lacking explicit stop signals, such as SPM for CodeGen and IPF for CodeLLaMa-34B. In these scenarios, our truncation method allows fair comparisons across various models by standardizing the completion endpoint.

%% file: tables/table_appendix_trunc_results_block.tex
\begin{table}[htb!]
\caption{Comparison of model performance with and without our syntax-aware truncation algorithm in the post-processing phase. This table presents two numbers for each model evaluated: \textbf{Pass@1} and \textbf{CErr\%}, as well as the prompt selected to evaluate each model.}
\label{tab:appendix_truncation_block}
\vskip 0.15in
\begin{center}
\scalebox{0.9}{
\begin{tabular}{lccccl}
\toprule
& \multicolumn{2}{c}{\textbf{No Trunc.}} & \multicolumn{2}{c}{\textbf{Syntax Trunc.}} & \\
\cmidrule(lr){2-3} \cmidrule(lr){4-5}
& \textbf{Pass@1} & \textbf{CErr\%} & \textbf{Pass@1} & \textbf{CErr\%} & \textbf{Prompt} \\ \midrule

GPT-3.5 (175B) & \colorbox[HTML]{FFF8F8}{28.7} & \colorbox[HTML]{FFE3E3}{25.3} & \colorbox[HTML]{FBFFFB}{31.2} & \colorbox[HTML]{FFEEEE}{17.0} & 1S \\
GPT-4 ($>$ 220B) & \colorbox[HTML]{D0FFD0}{41.7} & \colorbox[HTML]{FFE3E3}{25.4} & \colorbox[HTML]{CEFFCE}{42.1} & \colorbox[HTML]{FFE6E6}{22.9} & 1S \\
CodeGen-16B & \colorbox[HTML]{FF8080}{\hspace{5pt}0.0} & \colorbox[HTML]{FF8080}{99.9} & \colorbox[HTML]{FFECEC}{25.9} & \colorbox[HTML]{FFEDED}{17.9} & SPM \\
InCoder-6B & \colorbox[HTML]{FFDBDB}{21.8} & \colorbox[HTML]{FFE2E2}{25.7} & \colorbox[HTML]{FFE9E9}{25.2} & \colorbox[HTML]{FFF3F3}{13.2} & PSM \\
CodeLLaMa-13B & \colorbox[HTML]{FFC4C4}{16.4} & \colorbox[HTML]{FFAEAE}{64.6} & \colorbox[HTML]{D1FFD1}{41.4} & \colorbox[HTML]{FFF6F6}{10.9} & SPM \\
CodeLLaMa-34B & \colorbox[HTML]{FF8484}{\hspace{5pt}1.0} & \colorbox[HTML]{FF8787}{94.5} & \colorbox[HTML]{DDFFDD}{38.5} & \colorbox[HTML]{FFF1F1}{14.7} & SPM \\
StarCoder (15.5B) & \colorbox[HTML]{CBFFCB}{42.7} & \colorbox[HTML]{FFF1F1}{14.3} & \colorbox[HTML]{C5FFC5}{44.1} & \colorbox[HTML]{FFF8F8}{\hspace{5pt}9.5} & SPM \\
DeepSeekCoder-33B & \colorbox[HTML]{84FF84}{59.7} & \colorbox[HTML]{FFFAFA}{\hspace{5pt}8.0} & \colorbox[HTML]{80FF80}{60.8} & \colorbox[HTML]{FFFFFF}{\hspace{5pt}4.0} & PSM \\

\bottomrule
\end{tabular}
}
\end{center}
\vskip -0.1in
\end{table}

%% file: tables/table_appendix_trunc_results_control.tex
\begin{table}[htb!]
\caption{Comparison of model performance with and without our syntax-aware truncation algorithm in the post-processing phase on control-flow expression completion. This table presents two numbers for each model evaluated: \textbf{Pass@1} and \textbf{CErr\%}, as well as the prompt selected to evaluate each model.}
\label{tab:appendix_truncation_control}
\vskip 0.15in
\begin{center}
\scalebox{0.9}{
\begin{tabular}{lccccl}
\toprule
& \multicolumn{2}{c}{\textbf{No Trunc.}} & \multicolumn{2}{c}{\textbf{Syntax Trunc.}} & \\
\cmidrule(lr){2-3} \cmidrule(lr){4-5}
& \textbf{Pass@1} & \textbf{CErr\%} & \textbf{Pass@1} & \textbf{CErr\%} & \textbf{Prompt} \\ \midrule

GPT-3.5 (175B) & \colorbox[HTML]{F8FFF8}{37.4} & \colorbox[HTML]{FFE7E7}{19.7} & \colorbox[HTML]{F8FFF8}{37.5} & \colorbox[HTML]{FFE7E7}{19.5} & 1S \\
GPT-4 ($>$ 220B) & \colorbox[HTML]{B9FFB9}{55.2} & \colorbox[HTML]{FFE4E4}{21.8} & \colorbox[HTML]{B9FFB9}{55.2} & \colorbox[HTML]{FFE4E4}{21.9} & 1S \\
CodeGen-16B & \colorbox[HTML]{FF8080}{\hspace{5pt}0.0} & \colorbox[HTML]{FF8080}{99.9} & \colorbox[HTML]{FEFFFE}{35.7} & \colorbox[HTML]{FFEEEE}{14.6} & SPM \\
InCoder-6B & \colorbox[HTML]{FFA5A5}{10.4} & \colorbox[HTML]{FFB0B0}{62.1} & \colorbox[HTML]{FFE5E5}{28.2} & \colorbox[HTML]{FFF2F2}{11.0} & PSM \\
CodeLLaMa-13B & \colorbox[HTML]{FFE3E3}{27.8} & \colorbox[HTML]{FFBABA}{54.8} & \colorbox[HTML]{B1FFB1}{57.2} & \colorbox[HTML]{FFFEFE}{\hspace{5pt}2.3} & SPM \\
CodeLLaMa-34B & \colorbox[HTML]{FF8080}{\hspace{5pt}0.3} & \colorbox[HTML]{FF8181}{98.6} & \colorbox[HTML]{BDFFBD}{54.0} & \colorbox[HTML]{FFFDFD}{\hspace{5pt}2.7} & SPM \\
StarCoder (15.5B) & \colorbox[HTML]{C5FFC5}{51.8} & \colorbox[HTML]{FFF4F4}{\hspace{5pt}9.8} & \colorbox[HTML]{BBFFBB}{54.5} & \colorbox[HTML]{FFF9F9}{\hspace{5pt}6.0} & PSM \\
DeepSeekCoder-33B & \colorbox[HTML]{82FF82}{70.3} & \colorbox[HTML]{FFFDFD}{\hspace{5pt}2.8} & \colorbox[HTML]{80FF80}{71.1} & \colorbox[HTML]{FFFFFF}{\hspace{5pt}1.1} & PSM \\

\bottomrule
\end{tabular}
}
\end{center}
\vskip -0.1in
\end{table}

%% file: tables/table_appendix_trunc_results_api.tex
\begin{table}[htb!]
\caption{Comparison of model performance with and without our syntax-aware truncation algorithm in the post-processing phase on API function call completion. This table presents two numbers for each model evaluated: \textbf{Pass@1} and \textbf{CErr\%}, as well as the prompt selected to evaluate each model.}
\label{tab:appendix_truncation_api}
\vskip 0.15in
\begin{center}
\scalebox{0.9}{
\begin{tabular}{lccccl}
\toprule
& \multicolumn{2}{c}{\textbf{No Trunc.}} & \multicolumn{2}{c}{\textbf{Syntax Trunc.}} & \\
\cmidrule(lr){2-3} \cmidrule(lr){4-5}
& \textbf{Pass@1} & \textbf{CErr\%} & \textbf{Pass@1} & \textbf{CErr\%} & \textbf{Prompt} \\ \midrule

GPT-3.5 (175B) & \colorbox[HTML]{E9FFE9}{44.2} & \colorbox[HTML]{FFFFFF}{\hspace{5pt}0.0} & \colorbox[HTML]{C8FFC8}{53.9} & \colorbox[HTML]{FFFFFF}{\hspace{5pt}0.0} & 1S \\
GPT-4 ($>$ 220B) & \colorbox[HTML]{BCFFBC}{57.4} & \colorbox[HTML]{FFFFFF}{\hspace{5pt}0.0} & \colorbox[HTML]{AAFFAA}{62.6} & \colorbox[HTML]{FFFFFF}{\hspace{5pt}0.0} & 1S \\
CodeGen-16B & \colorbox[HTML]{FF8080}{\hspace{5pt}0.0} & \colorbox[HTML]{FF8080}{99.9} & \colorbox[HTML]{FFEAEA}{31.3} & \colorbox[HTML]{FFFDFD}{\hspace{5pt}1.9} & SPM \\
InCoder-6B & \colorbox[HTML]{FFD0D0}{23.9} & \colorbox[HTML]{FFFFFF}{\hspace{5pt}0.0} & \colorbox[HTML]{DBFFDB}{48.1} & \colorbox[HTML]{FFFFFF}{\hspace{5pt}0.0} & PSM \\
CodeLLaMa-13B & \colorbox[HTML]{FFF1F1}{33.5} & \colorbox[HTML]{FFFFFF}{\hspace{5pt}0.0} & \colorbox[HTML]{B4FFB4}{59.7} & \colorbox[HTML]{FFFFFF}{\hspace{5pt}0.0} & SPM \\
CodeLLaMa-34B & \colorbox[HTML]{FFA8A8}{11.9} & \colorbox[HTML]{FFFFFF}{\hspace{5pt}0.0} & \colorbox[HTML]{BFFFBF}{56.5} & \colorbox[HTML]{FFFEFE}{\hspace{5pt}0.6} & IPF \\
StarCoder (15.5B) & \colorbox[HTML]{9FFF9F}{65.8} & \colorbox[HTML]{FFFFFF}{\hspace{5pt}0.3} & \colorbox[HTML]{98FF98}{68.1} & \colorbox[HTML]{FFFFFF}{\hspace{5pt}0.3} & PSM \\
DeepSeekCoder-33B & \colorbox[HTML]{89FF89}{72.3} & \colorbox[HTML]{FFFFFF}{\hspace{5pt}0.0} & \colorbox[HTML]{80FF80}{75.2} & \colorbox[HTML]{FFFFFF}{\hspace{5pt}0.0} & PSM \\

\bottomrule
\end{tabular}
}
\end{center}
\vskip -0.1in
\end{table}

%% file: contents/appendix/additional_main_results.tex
\subsection{Evaluation Results of More LLMs}\label{sec:additional_main_results}

In this section, we expand our evaluation to include additional LLMs: Mixtral~\citep{mixtral}, Phi~\citep{phi}, WizardCoder~\citep{wizardcoder}, and Magicoder~\citep{magicoder}.

\input{tables/table_appendix_additional_models}

\Cref{tab:appendix_additional_models} provides the details of the additional models. Mixtral-8x7B, a sparse mixture-of-experts (MoE) model, uses a router to select two expert 7B models for each inference. The Phi models are small LLMs pretrained using distilled data (synthetic data generated by a teacher LLM). WizardCoder and Magicoder are initialized with base models and then finetuned on distilled data. Specifically, WizardCoder variants (15B and 33B) use StarCoder and DeepSeekCoder-33B as their respective base models, while Magicoder-S-DS-6.7B is finetuned from DeepSeekCoder-6.7B. Notably, these models are classified as FIM models because their inherited vocabulary supports FIM special tokens, despite the finetuning process not directly engaging with FIM tasks.

\input{tables/table_appendix_main}

\Cref{tab:appendix_main} shows our experimental results, which yield the following insights:

\begin{itemize}
\item Mixtral-8x7B achieves performance comparable to CodeLLaMa-7B. Given that Mixtral is not specialized for coding, its comparable performance to 7B code LLMs shows the effectiveness of MoE. Typically, general-purpose LLMs like GPT-3.5, GPT-4, and Mixtral need more parameters to match the performance of specialized code LLMs.
\item Models Pretrained on Distilled Data (Phi-1.5 and Phi-2) exhibit good performance considering their tiny sizes, but they don’t reach the high standards set by their HumanEval results. This difference underscores the \ours benchmark's diversity and complexity compared to HumanEval.
\item Models Finetuned on Distilled Data shows a slight drop in performance compared to their FIM-pretrained base models (e.g., WizardCoder-15B vs. StarCoder, WizardCoder-33B vs. DeepSeekCoder-33B, Magicoder-S-DS-6.7B vs. DeepSeekCoder-6.7B). The performance drop stems from the left-to-right finetuning on distilled data, which lacks FIM objectives, thereby harming the models' proficiency in FIM tasks.
\end{itemize}

These additional findings further reinforce our original conclusion: the pretraining methodology significantly influences the performance of code LLMs.

%% file: tables/table_appendix_additional_models.tex
\begin{table}[tb]
\caption{Summary of evaluated models, highlighting data cutoff dates, open-source status (OS), and pretraining objectives. Dates in \textcolor{red}{red} indicate overlap between the model's pretraining data and the \ours benchmark in date range (post-April 2022). Data cutoff dates for InCoder are estimated based on their initial paper draft publication dates. The OS column denotes open-source availability ($\surd$ for yes, $\times$ for no), and the FIM column indicates models pretrained with FIM objectives and support for sentinel tokens in FIM inference.}
\label{tab:appendix_additional_models}
\vskip 0.15in
\begin{center}
\scalebox{0.9}{
\begin{tabular}{lllcc}
\toprule
               & \textbf{\#Params} & \textbf{Data Cutoff}            & \textbf{OS} & \textbf{FIM} \\ \midrule
GPT-4          & -                 & Sept 2021                       & $\times$    & $\times$     \\
CodeGen        & 350M/2B/6B/16B    & Oct 2021                        & $\surd$     & $\times$     \\
InCoder        & 1.3B/6.7B         & $\le$ Mar 2022                  & $\surd$     & $\surd$      \\
CodeLLaMa      & 7B/13B            & \textcolor{red}{Jul 2022}       & $\surd$     & $\surd$      \\
CodeLLaMa      & 34B               & \textcolor{red}{Jul 2022}       & $\surd$     & $\times$     \\
StarCoder      & 15.5B             & Mar 2022                        & $\surd$     & $\surd$      \\
DeepSeekCoder  & 1.3B/6.7B/33B     & \textcolor{red}{Feb 2023}       & $\surd$     & $\surd$      \\ \midrule
Mixtral        & 46.7B (8x7B)      & \textcolor{red}{$\le$ Sep 2023} & $\surd$     & $\times$     \\
Phi            & 1.3B/2.7B         & Mar 2022                        & $\surd$     & $\times$     \\
WizardCoder    & 1B/3B/15B         & Mar 2022                        & $\surd$     & $\surd$  \\
WizardCoder    & 33B               & \textcolor{red}{Feb 2023}       & $\surd$     & $\surd$  \\
Magicoder      & 6.7B              & \textcolor{red}{Feb 2023}       & $\surd$     & $\surd$      \\\bottomrule
\end{tabular}
}
\end{center}
\vskip -0.1in
\end{table}

%% file: tables/table_appendix_main.tex
\begin{table}[h]
\caption{Pass@1 of various models on the \ours benchmark, showing their performance in algorithmic block completion (Algo.), control-flow completion (Control), and API function call completion (API).}
\label{tab:appendix_main}
\vskip 0.15in
\begin{center}
\scalebox{0.9}{
\begin{tabular}{lcccc}
\toprule
& \textbf{Algo.} & \textbf{Control} & \textbf{API} & \textbf{Avg}
\\ \midrule

GPT-3.5 (175B) & \colorbox[HTML]{FFD5D5}{31.2} & \colorbox[HTML]{FFCDCD}{37.5} & \colorbox[HTML]{E6FFE6}{53.9} & \colorbox[HTML]{FFE3E3}{40.9} \\
GPT-4 ($>$ 220B) & \colorbox[HTML]{EBFFEB}{42.1} & \colorbox[HTML]{D4FFD4}{55.2} & \colorbox[HTML]{BCFFBC}{62.6} & \colorbox[HTML]{D7FFD7}{53.3} \\
CodeGen-350M & \colorbox[HTML]{FF8080}{16.3} & \colorbox[HTML]{FF9090}{26.1} & \colorbox[HTML]{FF9494}{26.5} & \colorbox[HTML]{FF8080}{22.9} \\
CodeGen-2B & \colorbox[HTML]{FFA9A9}{23.5} & \colorbox[HTML]{FFB4B4}{32.9} & \colorbox[HTML]{FFB0B0}{32.3} & \colorbox[HTML]{FFA4A4}{29.5} \\
CodeGen-6B & \colorbox[HTML]{FFA9A9}{23.6} & \colorbox[HTML]{FFBFBF}{34.8} & \colorbox[HTML]{FF9A9A}{27.7} & \colorbox[HTML]{FF9F9F}{28.7} \\
CodeGen-16B & \colorbox[HTML]{FFB7B7}{25.9} & \colorbox[HTML]{FFC3C3}{35.7} & \colorbox[HTML]{FFABAB}{31.3} & \colorbox[HTML]{FFACAC}{31.0} \\
InCoder-1B & \colorbox[HTML]{FF9B9B}{21.1} & \colorbox[HTML]{FF8080}{22.9} & \colorbox[HTML]{FFE8E8}{43.9} & \colorbox[HTML]{FFA3A3}{29.3} \\
InCoder-6B & \colorbox[HTML]{FFB2B2}{25.2} & \colorbox[HTML]{FF9B9B}{28.2} & \colorbox[HTML]{FFFCFC}{48.1} & \colorbox[HTML]{FFBCBC}{33.8} \\
CodeLLaMa-7B & \colorbox[HTML]{FFE9E9}{34.7} & \colorbox[HTML]{DCFFDC}{53.6} & \colorbox[HTML]{FFF6F6}{46.8} & \colorbox[HTML]{FFFAFA}{45.0} \\
CodeLLaMa-13B & \colorbox[HTML]{EFFFEF}{41.4} & \colorbox[HTML]{C9FFC9}{57.2} & \colorbox[HTML]{CAFFCA}{59.7} & \colorbox[HTML]{D9FFD9}{52.8} \\
CodeLLaMa-34B & \colorbox[HTML]{FFFFFF}{38.5} & \colorbox[HTML]{DAFFDA}{54.0} & \colorbox[HTML]{DAFFDA}{56.5} & \colorbox[HTML]{EBFFEB}{49.7} \\
StarCoder (15.5B) & \colorbox[HTML]{DFFFDF}{44.1} & \colorbox[HTML]{D8FFD8}{54.5} & \colorbox[HTML]{A2FFA2}{68.1} & \colorbox[HTML]{CAFFCA}{55.5} \\
DeepSeekCoder-1.3B & \colorbox[HTML]{F0FFF0}{41.2} & \colorbox[HTML]{D9FFD9}{54.1} & \colorbox[HTML]{BCFFBC}{62.6} & \colorbox[HTML]{DAFFDA}{52.6} \\
DeepSeekCoder-6.7B & \colorbox[HTML]{A2FFA2}{54.7} & \colorbox[HTML]{9CFF9C}{65.8} & \colorbox[HTML]{9AFF9A}{69.7} & \colorbox[HTML]{9FFF9F}{63.4} \\
DeepSeekCoder-33B & \colorbox[HTML]{80FF80}{\textbf{60.8}} & \colorbox[HTML]{80FF80}{\textbf{71.1}} & \colorbox[HTML]{80FF80}{\textbf{75.2}} & \colorbox[HTML]{80FF80}{\textbf{69.0}} \\ \midrule
Mixtral-8x7B & \colorbox[HTML]{FFE3E3}{33.7} & \colorbox[HTML]{EEFFEE}{50.3} & \colorbox[HTML]{D0FFD0}{58.4} & \colorbox[HTML]{F7FFF7}{47.5} \\
Phi-1.5 (1.3B) & \colorbox[HTML]{FF8F8F}{19.0} & \colorbox[HTML]{FFA5A5}{29.9} & \colorbox[HTML]{FF9A9A}{27.7} & \colorbox[HTML]{FF8E8E}{25.5} \\
Phi-2 (2.7B) & \colorbox[HTML]{FFAAAA}{23.8} & \colorbox[HTML]{FFBEBE}{34.8} & \colorbox[HTML]{FF8080}{22.3} & \colorbox[HTML]{FF9696}{26.9} \\
WizardCoder-1B & \colorbox[HTML]{FFC3C3}{28.1} & \colorbox[HTML]{FFDADA}{40.0} & \colorbox[HTML]{D5FFD5}{57.4} & \colorbox[HTML]{FFE8E8}{41.8} \\
WizardCoder-3B & \colorbox[HTML]{FFE7E7}{34.4} & \colorbox[HTML]{FFFBFB}{46.3} & \colorbox[HTML]{B0FFB0}{65.2} & \colorbox[HTML]{F0FFF0}{48.6} \\
WizardCoder-15B & \colorbox[HTML]{F1FFF1}{41.0} & \colorbox[HTML]{E2FFE2}{52.6} & \colorbox[HTML]{94FF94}{71.0} & \colorbox[HTML]{CEFFCE}{54.8} \\
WizardCoder-33B & \colorbox[HTML]{C0FFC0}{49.5} & \colorbox[HTML]{99FF99}{66.3} & \colorbox[HTML]{83FF83}{74.5} & \colorbox[HTML]{9EFF9E}{63.4} \\
Magicoder-S-DS-6.7B & \colorbox[HTML]{EEFFEE}{41.5} & \colorbox[HTML]{AEFFAE}{62.3} & \colorbox[HTML]{AEFFAE}{65.5} & \colorbox[HTML]{C5FFC5}{56.4} \\

\bottomrule
\end{tabular}
}
\end{center}
\vskip -0.1in
\end{table}

%% file: contents/appendix/per_pl_results.tex
\subsection{Result Analysis by Programming Languages}\label{sec:appendix_per_pl_results}

\input{tables/table_appendix_per_pl_results}

Table \ref{tab:table_appendix_per_pl_results} shows the average pass@1 rate for each LLM in our \ours benchmark, broken down by programming language and averaged on three completion tasks. Our analysis reveals that:

\paragraph{LLMs exhibit higher success rates in Java and C\#,} likely due to the verbosity of these languages, which leads to more predictable coding patterns. Conversely, completion in C++ and Python is more challenging, due to the more concise and less predictable coding styles prevalent among developers. As we discussed in \Cref{sec:benchmark_statistics}, the \ours benchmark consist of different programming languages written by different developers, so the results are affected by intrinsic variability in coding styles across PLs.

\paragraph{Despite the language-dependent variability, the relative rankings of LLMs stay mostly consistent.} This underscores the robustness of the \ours benchmark and supports our decision to report \textit{micro-averaged} performance metrics in our study.

%% file: tables/table_appendix_per_pl_results.tex
\begin{table}[h]
\caption{Average pass@1 of various models on the three tasks in \ours, showing their results in different programming languages.}
\label{tab:table_appendix_per_pl_results}
\vskip 0.15in
\begin{center}
\scalebox{0.9}{
\begin{tabular}{lccccc}
\toprule
& \textbf{C++} & \textbf{Java} & \textbf{Python} & \textbf{C\#} & \textbf{Avg}
\\ \midrule

GPT-3.5 (175B) & \colorbox[HTML]{FFEFEF}{39.3} & \colorbox[HTML]{F6FFF6}{54.2} & \colorbox[HTML]{FFB3B3}{29.5} & \colorbox[HTML]{FFE2E2}{40.5} & \colorbox[HTML]{FFE3E3}{40.9} \\
GPT-4 ($>$ 220B) & \colorbox[HTML]{D8FFD8}{49.4} & \colorbox[HTML]{C2FFC2}{63.3} & \colorbox[HTML]{FFF1F1}{42.7} & \colorbox[HTML]{DBFFDB}{54.6} & \colorbox[HTML]{D7FFD7}{53.3} \\
CodeGen-350M & \colorbox[HTML]{FF9797}{23.1} & \colorbox[HTML]{FF9393}{33.6} & \colorbox[HTML]{FF8080}{18.7} & \colorbox[HTML]{FF8484}{19.9} & \colorbox[HTML]{FF8080}{22.9} \\
CodeGen-2B & \colorbox[HTML]{FFB1B1}{27.9} & \colorbox[HTML]{FFCBCB}{43.4} & \colorbox[HTML]{FF9999}{24.1} & \colorbox[HTML]{FFADAD}{28.9} & \colorbox[HTML]{FFA4A4}{29.5} \\
CodeGen-6B & \colorbox[HTML]{FFBEBE}{30.3} & \colorbox[HTML]{FFD2D2}{44.6} & \colorbox[HTML]{FF8C8C}{21.2} & \colorbox[HTML]{FFA2A2}{26.4} & \colorbox[HTML]{FF9F9F}{28.7} \\
CodeGen-16B & \colorbox[HTML]{FFDADA}{35.5} & \colorbox[HTML]{FFDCDC}{46.5} & \colorbox[HTML]{FF8989}{20.7} & \colorbox[HTML]{FFB3B3}{30.2} & \colorbox[HTML]{FFACAC}{31.0} \\
InCoder-1B & \colorbox[HTML]{FF8D8D}{21.3} & \colorbox[HTML]{FFA0A0}{35.9} & \colorbox[HTML]{FFCECE}{35.3} & \colorbox[HTML]{FFBCBC}{32.2} & \colorbox[HTML]{FFA3A3}{29.3} \\
InCoder-6B & \colorbox[HTML]{FFA8A8}{26.2} & \colorbox[HTML]{FFC0C0}{41.4} & \colorbox[HTML]{FFE7E7}{40.5} & \colorbox[HTML]{FFBDBD}{32.4} & \colorbox[HTML]{FFBCBC}{33.8} \\
CodeLLaMa-7B & \colorbox[HTML]{FFD0D0}{33.6} & \colorbox[HTML]{EBFFEB}{56.1} & \colorbox[HTML]{FFE7E7}{40.6} & \colorbox[HTML]{FAFFFA}{47.9} & \colorbox[HTML]{FFFAFA}{45.0} \\
CodeLLaMa-13B & \colorbox[HTML]{ECFFEC}{45.8} & \colorbox[HTML]{D4FFD4}{60.2} & \colorbox[HTML]{DEFFDE}{52.5} & \colorbox[HTML]{AEFFAE}{64.7} & \colorbox[HTML]{D9FFD9}{52.8} \\
CodeLLaMa-34B & \colorbox[HTML]{FAFFFA}{43.3} & \colorbox[HTML]{D5FFD5}{59.9} & \colorbox[HTML]{EFFFEF}{49.0} & \colorbox[HTML]{B9FFB9}{62.3} & \colorbox[HTML]{EBFFEB}{49.7} \\
StarCoder (15.5B) & \colorbox[HTML]{CAFFCA}{52.0} & \colorbox[HTML]{BFFFBF}{63.9} & \colorbox[HTML]{BDFFBD}{59.5} & \colorbox[HTML]{DBFFDB}{54.7} & \colorbox[HTML]{CAFFCA}{55.5} \\
DeepSeekCoder-1.3B & \colorbox[HTML]{F2FFF2}{44.7} & \colorbox[HTML]{CEFFCE}{61.3} & \colorbox[HTML]{C6FFC6}{57.7} & \colorbox[HTML]{D8FFD8}{55.5} & \colorbox[HTML]{DAFFDA}{52.6} \\
DeepSeekCoder-6.7B & \colorbox[HTML]{ACFFAC}{57.6} & \colorbox[HTML]{9BFF9B}{70.3} & \colorbox[HTML]{97FF97}{67.5} & \colorbox[HTML]{93FF93}{70.4} & \colorbox[HTML]{9FFF9F}{63.4} \\
DeepSeekCoder-33B & \colorbox[HTML]{80FF80}{65.8} & \colorbox[HTML]{80FF80}{75.1} & \colorbox[HTML]{80FF80}{72.5} & \colorbox[HTML]{80FF80}{74.7} & \colorbox[HTML]{80FF80}{69.0} \\
Mixtral-8x7B & \colorbox[HTML]{FFFDFD}{42.0} & \colorbox[HTML]{E1FFE1}{58.0} & \colorbox[HTML]{FFF3F3}{43.0} & \colorbox[HTML]{D1FFD1}{56.9} & \colorbox[HTML]{F7FFF7}{47.5} \\
Phi-1.5 (1.3B) & \colorbox[HTML]{FF8080}{18.8} & \colorbox[HTML]{FF8080}{30.1} & \colorbox[HTML]{FFA5A5}{26.7} & \colorbox[HTML]{FF8080}{18.9} & \colorbox[HTML]{FF8E8E}{25.5} \\
Phi-2 (2.7B) & \colorbox[HTML]{FF9494}{22.6} & \colorbox[HTML]{FFA3A3}{36.3} & \colorbox[HTML]{FF9898}{23.8} & \colorbox[HTML]{FF9393}{23.3} & \colorbox[HTML]{FF9696}{26.9} \\
WizardCoder-1B & \colorbox[HTML]{FFD3D3}{34.1} & \colorbox[HTML]{FFE9E9}{48.7} & \colorbox[HTML]{FFFAFA}{44.5} & \colorbox[HTML]{FFDCDC}{39.3} & \colorbox[HTML]{FFE8E8}{41.8} \\
WizardCoder-3B & \colorbox[HTML]{FAFFFA}{43.3} & \colorbox[HTML]{E6FFE6}{57.0} & \colorbox[HTML]{D8FFD8}{53.8} & \colorbox[HTML]{F2FFF2}{49.8} & \colorbox[HTML]{F0FFF0}{48.6} \\
WizardCoder-15B & \colorbox[HTML]{CCFFCC}{51.7} & \colorbox[HTML]{CBFFCB}{61.7} & \colorbox[HTML]{BFFFBF}{59.2} & \colorbox[HTML]{E7FFE7}{52.1} & \colorbox[HTML]{CEFFCE}{54.8} \\
WizardCoder-33B & \colorbox[HTML]{99FF99}{61.1} & \colorbox[HTML]{98FF98}{70.8} & \colorbox[HTML]{99FF99}{67.2} & \colorbox[HTML]{DAFFDA}{55.0} & \colorbox[HTML]{9EFF9E}{63.4} \\
Magicoder-S-DS-6.7B & \colorbox[HTML]{D1FFD1}{50.8} & \colorbox[HTML]{B6FFB6}{65.4} & \colorbox[HTML]{C8FFC8}{57.3} & \colorbox[HTML]{BCFFBC}{61.4} & \colorbox[HTML]{C5FFC5}{56.4} \\

\bottomrule
\end{tabular}
}
\end{center}
\vskip -0.1in
\end{table}

%% file: contents/appendix/case_study.tex
\subsection{Case Study}\label{sec:case_study}

This section presents a case study of the algorithmic block completion task from \ours (\texttt{task\_id: block\_completion\_008121}). Two similar-performing models, InCoder-6B and CodeCen-16B (achieving pass rates at 25.2\% and 25.9\% respectively), are compared.

The case originates from problem 1678B1 on Codeforces\footnote{\url{https://codeforces.com/problemset/problem/1678/B1}}; given the problem description, the task is to fill in the \texttt{\# TODO: Your code here} part of the provided code:

\begin{minted}{python}
t=int(input(""))
for z in range(t):
    n=int(input(""))
    a=input("")
    s=[]
    for i in range(0,len(a)-1,2):
        # TODO: Your code here
    b=s.count('10')
    c=s.count('01')
    print(b+c)
\end{minted}

The ground truth involves appending two characters from the string \texttt{a} to the list \texttt{s} in each loop iteration (we added 8 spaces at the beginning for clarity):

\begin{minted}{python}
        ab=a[i]+a[i+1]
        s.append(ab)
\end{minted}

Using the PSM prompt, InCoder-6B successfully generates a valid Python completion using Python string slicing:

\begin{minted}{python}
        s.append(a[i:i+2])
\end{minted}

CodeGen-16B uses the SPM prompt. Note that CodeGen-16B lacks EOS token support, leading to generation of extra code followed by infinite output generation unless truncated:

\begin{minted}{python}
        s.append(a[i])
    s.append(a[len(a)-1])
    s=s[::-1]
    b=s.count('10')
    c=s.count('01')
    print(b+c)

# [infinite empty lines]
\end{minted}

Applying syntax-aware truncation, we keep only the relevant block completion \texttt{s.append(a[i])}. Unfortunately, this still yields an incorrect solution, leading to an outcome of ``wrong answer'' in evaluation.

This case study underscores the significance of syntax-aware truncation and highlights the behavior of different models.

%% file: contents/appendix/data_contamination.tex
\subsection{Further Analysis on Data Contamination}\label{sec:appendix_data_contamination}

\ours is sourced from Codeforces contests and Github code commits created between April 1, 2022 and January 1, 2023. This period, unfortunately, overlaps with the pretraining data of CodeLLaMa and DeepSeekCoder. To analyze the potential influence of data contamination on our evaluation results, we create an new dataset for the algorithmic block completion task based on Codeforces contests from April 1, 2023, to January 31, 2024, without any overlap with the training data of these models. Then we evaluate each of these models and StarCoder, on both datasets. The findings are shown in \Cref{tab:appendix_data_contamination}. We also visualize each model's performance across various months in the new test dataset in \Cref{fig:data_contamination}.

\input{tables/table_appendix_data_contamination}

\input{figures/data_contamination}

Based on \Cref{tab:appendix_data_contamination}, for the new test data, without any overlap with the models' training date ranges, no significant performance decrease is noticed compared to the original dataset, which had a date range overlap. \Cref{fig:data_contamination} also shows stable performance across the timeline for all models, without a noticeable decline on newer questions for CodeLLaMa or DeepSeekCoder. These findings suggest that while vigilance against data contamination is prudent, the difference in cutoff dates has a negligible impact on our current evaluation results.

%% file: tables/table_appendix_data_contamination.tex
\begin{table}[h]
\caption{Pass@1 of each model on two versions of algorithmic block completion, including the \textbf{original} version (Apr 2022 - Jan 2023) and the \textbf{new} version (Apr 2023 - Jan 2024). Numbers in \textcolor{red}{red} indicate overlap between the model's pretraining data and the test dataset in date range. The $\Delta$ column shows the pass@1 change between the original and the new test datasets.}
\label{tab:appendix_data_contamination}
\vskip 0.15in
\begin{center}
\scalebox{0.9}{
\begin{tabular}{llccc}
\toprule
& \textbf{Data Cutoff} & \textbf{Original} & \textbf{New} & $\Delta$
\\ \midrule

StarCoder & Mar 2022 & 44.1 & 46.7 & +2.56 \\
CodeLLaMa-7B & Jul 2022 & \textcolor{red}{34.7} & 32.7 & -1.95 \\
CodeLLaMa-13B & Jul 2022 & \textcolor{red}{41.4} & 45.8 & +4.40 \\
CodeLLaMa-34B & Jul 2022 & \textcolor{red}{38.5} & 43.8 & +5.29 \\
DeepSeekCoder-1.3B & Feb 2023 & \textcolor{red}{41.2} & 46.1 & +4.87 \\
DeepSeekCoder-6.7B & Feb 2023 & \textcolor{red}{54.7} & 58.4 & +3.65 \\
DeepSeekCoder-33B & Feb 2023 & \textcolor{red}{60.8} & 61.7 & +0.91 \\

\bottomrule
\end{tabular}
}
\end{center}
\vskip -0.1in
\end{table}

%% file: figures/data_contamination.tex
\begin{figure}[htb!]
\vskip 0.2in
\begin{center}
\centerline{\includegraphics{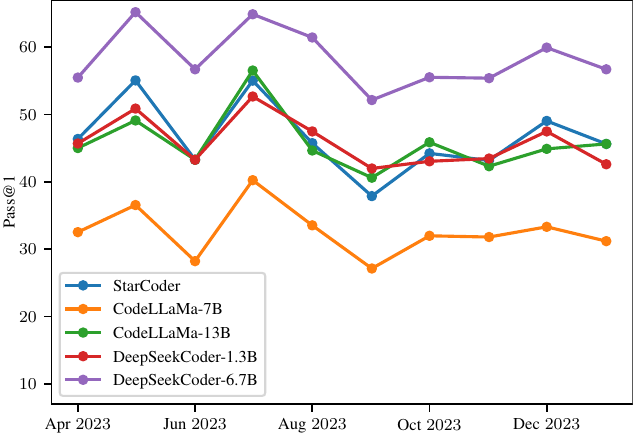}}
\caption{Pass@1 scores for each model on algorithmic block completion across various months in the \textbf{new} test dataset.}
\label{fig:data_contamination}
\end{center}
\vskip -0.2in
\end{figure}

%% file: ms.bbl
\begin{thebibliography}{51}
\providecommand{\natexlab}[1]{#1}
\providecommand{\url}[1]{\texttt{#1}}
\expandafter\ifx\csname urlstyle\endcsname\relax
  \providecommand{\doi}[1]{doi: #1}\else
  \providecommand{\doi}{doi: \begingroup \urlstyle{rm}\Url}\fi

\bibitem[Agashe et~al.(2019)Agashe, Iyer, and Zettlemoyer]{juice}
Agashe, R., Iyer, S., and Zettlemoyer, L.
\newblock {J}u{IC}e: A large scale distantly supervised dataset for open domain context-based code generation.
\newblock \penalty0 (arXiv:1910.02216), October 2019.
\newblock \doi{10.48550/arXiv.1910.02216}.
\newblock URL \url{http://arxiv.org/abs/1910.02216}.
\newblock arXiv:1910.02216 [cs].

\bibitem[Ahmad et~al.(2021)Ahmad, Chakraborty, Ray, and Chang]{plbart}
Ahmad, W.~U., Chakraborty, S., Ray, B., and Chang, K.-W.
\newblock Unified pre-training for program understanding and generation.
\newblock Apr 2021.
\newblock \doi{10.48550/arXiv.2103.06333}.
\newblock URL \url{http://arxiv.org/abs/2103.06333}.
\newblock arXiv:2103.06333 [cs].

\bibitem[Allal et~al.(2023)Allal, Li, Kocetkov, Mou, Akiki, Ferrandis, Muennighoff, Mishra, Gu, Dey, Umapathi, Anderson, Zi, Poirier, Schoelkopf, Troshin, Abulkhanov, Romero, Lappert, De~Toni, del Río, Liu, Bose, Bhattacharyya, Zhuo, Yu, Villegas, Zocca, Mangrulkar, Lansky, Nguyen, Contractor, Villa, Li, Bahdanau, Jernite, Hughes, Fried, Guha, de~Vries, and von Werra]{santacoder}
Allal, L.~B., Li, R., Kocetkov, D., Mou, C., Akiki, C., Ferrandis, C.~M., Muennighoff, N., Mishra, M., Gu, A., Dey, M., Umapathi, L.~K., Anderson, C.~J., Zi, Y., Poirier, J.~L., Schoelkopf, H., Troshin, S., Abulkhanov, D., Romero, M., Lappert, M., De~Toni, F., del Río, B.~G., Liu, Q., Bose, S., Bhattacharyya, U., Zhuo, T.~Y., Yu, I., Villegas, P., Zocca, M., Mangrulkar, S., Lansky, D., Nguyen, H., Contractor, D., Villa, L., Li, J., Bahdanau, D., Jernite, Y., Hughes, S., Fried, D., Guha, A., de~Vries, H., and von Werra, L.
\newblock {S}anta{C}oder: don’t reach for the stars!
\newblock \penalty0 (arXiv:2301.03988), February 2023.
\newblock \doi{10.48550/arXiv.2301.03988}.
\newblock URL \url{http://arxiv.org/abs/2301.03988}.
\newblock arXiv:2301.03988 [cs].

\bibitem[Athiwaratkun et~al.(2023)Athiwaratkun, Gouda, Wang, Li, Tian, Tan, Ahmad, Wang, Sun, Shang, Gonugondla, Ding, Kumar, Fulton, Farahani, Jain, Giaquinto, Qian, Ramanathan, Nallapati, Ray, Bhatia, Sengupta, Roth, and Xiang]{mbxp}
Athiwaratkun, B., Gouda, S.~K., Wang, Z., Li, X., Tian, Y., Tan, M., Ahmad, W.~U., Wang, S., Sun, Q., Shang, M., Gonugondla, S.~K., Ding, H., Kumar, V., Fulton, N., Farahani, A., Jain, S., Giaquinto, R., Qian, H., Ramanathan, M.~K., Nallapati, R., Ray, B., Bhatia, P., Sengupta, S., Roth, D., and Xiang, B.
\newblock Multi-lingual evaluation of code generation models.
\newblock \penalty0 (arXiv:2210.14868), March 2023.
\newblock \doi{10.48550/arXiv.2210.14868}.
\newblock URL \url{http://arxiv.org/abs/2210.14868}.
\newblock arXiv:2210.14868 [cs].

\bibitem[Austin et~al.(2021)Austin, Odena, Nye, Bosma, Michalewski, Dohan, Jiang, Cai, Terry, Le, and Sutton]{mbpp}
Austin, J., Odena, A., Nye, M., Bosma, M., Michalewski, H., Dohan, D., Jiang, E., Cai, C., Terry, M., Le, Q., and Sutton, C.
\newblock Program synthesis with large language models.
\newblock Aug 2021.
\newblock \doi{10.48550/arXiv.2108.07732}.
\newblock URL \url{http://arxiv.org/abs/2108.07732}.
\newblock arXiv:2108.07732 [cs].

\bibitem[Bavarian et~al.(2022)Bavarian, Jun, Tezak, Schulman, McLeavey, Tworek, and Chen]{humaneval_fim}
Bavarian, M., Jun, H., Tezak, N., Schulman, J., McLeavey, C., Tworek, J., and Chen, M.
\newblock Efficient training of language models to fill in the middle.
\newblock \penalty0 (arXiv:2207.14255), July 2022.
\newblock \doi{10.48550/arXiv.2207.14255}.
\newblock URL \url{http://arxiv.org/abs/2207.14255}.
\newblock arXiv:2207.14255 [cs].

\bibitem[Brown et~al.(2020)Brown, Mann, Ryder, Subbiah, Kaplan, Dhariwal, Neelakantan, Shyam, Sastry, Askell, Agarwal, Herbert-Voss, Krueger, Henighan, Child, Ramesh, Ziegler, Wu, Winter, Hesse, Chen, Sigler, Litwin, Gray, Chess, Clark, Berner, McCandlish, Radford, Sutskever, and Amodei]{gpt3}
Brown, T.~B., Mann, B., Ryder, N., Subbiah, M., Kaplan, J., Dhariwal, P., Neelakantan, A., Shyam, P., Sastry, G., Askell, A., Agarwal, S., Herbert-Voss, A., Krueger, G., Henighan, T., Child, R., Ramesh, A., Ziegler, D.~M., Wu, J., Winter, C., Hesse, C., Chen, M., Sigler, E., Litwin, M., Gray, S., Chess, B., Clark, J., Berner, C., McCandlish, S., Radford, A., Sutskever, I., and Amodei, D.
\newblock Language models are few-shot learners.
\newblock \penalty0 (arXiv:2005.14165), July 2020.
\newblock \doi{10.48550/arXiv.2005.14165}.
\newblock URL \url{http://arxiv.org/abs/2005.14165}.
\newblock arXiv:2005.14165 [cs].

\bibitem[Cassano et~al.(2022)Cassano, Gouwar, Nguyen, Nguyen, Phipps-Costin, Pinckney, Yee, Zi, Anderson, Feldman, Guha, Greenberg, and Jangda]{multiple}
Cassano, F., Gouwar, J., Nguyen, D., Nguyen, S., Phipps-Costin, L., Pinckney, D., Yee, M.-H., Zi, Y., Anderson, C.~J., Feldman, M.~Q., Guha, A., Greenberg, M., and Jangda, A.
\newblock {M}ulti{PL}-{E}: A scalable and extensible approach to benchmarking neural code generation.
\newblock \penalty0 (arXiv:2208.08227), December 2022.
\newblock \doi{10.48550/arXiv.2208.08227}.
\newblock URL \url{http://arxiv.org/abs/2208.08227}.
\newblock arXiv:2208.08227 [cs].

\bibitem[Chen et~al.(2021{\natexlab{a}})Chen, Tworek, Jun, Yuan, Pinto, Kaplan, Edwards, Burda, Joseph, Brockman, Ray, Puri, Krueger, Petrov, Khlaaf, Sastry, Mishkin, Chan, Gray, Ryder, Pavlov, Power, Kaiser, Bavarian, Winter, Tillet, Such, Cummings, Plappert, Chantzis, Barnes, Herbert-Voss, Guss, Nichol, Paino, Tezak, Tang, Babuschkin, Balaji, Jain, Saunders, Hesse, Carr, Leike, Achiam, Misra, Morikawa, Radford, Knight, Brundage, Murati, Mayer, Welinder, McGrew, Amodei, McCandlish, Sutskever, and Zaremba]{codex}
Chen, M., Tworek, J., Jun, H., Yuan, Q., Pinto, H. P. d.~O., Kaplan, J., Edwards, H., Burda, Y., Joseph, N., Brockman, G., Ray, A., Puri, R., Krueger, G., Petrov, M., Khlaaf, H., Sastry, G., Mishkin, P., Chan, B., Gray, S., Ryder, N., Pavlov, M., Power, A., Kaiser, L., Bavarian, M., Winter, C., Tillet, P., Such, F.~P., Cummings, D., Plappert, M., Chantzis, F., Barnes, E., Herbert-Voss, A., Guss, W.~H., Nichol, A., Paino, A., Tezak, N., Tang, J., Babuschkin, I., Balaji, S., Jain, S., Saunders, W., Hesse, C., Carr, A.~N., Leike, J., Achiam, J., Misra, V., Morikawa, E., Radford, A., Knight, M., Brundage, M., Murati, M., Mayer, K., Welinder, P., McGrew, B., Amodei, D., McCandlish, S., Sutskever, I., and Zaremba, W.
\newblock Evaluating large language models trained on code.
\newblock Jul 2021{\natexlab{a}}.
\newblock \doi{10.48550/arXiv.2107.03374}.
\newblock URL \url{http://arxiv.org/abs/2107.03374}.
\newblock arXiv:2107.03374 [cs].

\bibitem[Chen et~al.(2021{\natexlab{b}})Chen, Gong, Cheung, and Song]{plotcoder}
Chen, X., Gong, L., Cheung, A., and Song, D.
\newblock {P}lot{C}oder: Hierarchical decoding for synthesizing visualization code in programmatic context.
\newblock In Zong, C., Xia, F., Li, W., and Navigli, R. (eds.), \emph{Proceedings of the 59th Annual Meeting of the Association for Computational Linguistics and the 11th International Joint Conference on Natural Language Processing (Volume 1: Long Papers)}, pp.\  2169--2181, Online, August 2021{\natexlab{b}}. Association for Computational Linguistics.
\newblock \doi{10.18653/v1/2021.acl-long.169}.
\newblock URL \url{https://aclanthology.org/2021.acl-long.169}.

\bibitem[Chowdhery et~al.(2022)Chowdhery, Narang, Devlin, Bosma, Mishra, Roberts, Barham, Chung, Sutton, Gehrmann, Schuh, Shi, Tsvyashchenko, Maynez, Rao, Barnes, Tay, Shazeer, Prabhakaran, Reif, Du, Hutchinson, Pope, Bradbury, Austin, Isard, Gur-Ari, Yin, Duke, Levskaya, Ghemawat, Dev, Michalewski, Garcia, Misra, Robinson, Fedus, Zhou, Ippolito, Luan, Lim, Zoph, Spiridonov, Sepassi, Dohan, Agrawal, Omernick, Dai, Pillai, Pellat, Lewkowycz, Moreira, Child, Polozov, Lee, Zhou, Wang, Saeta, Diaz, Firat, Catasta, Wei, Meier-Hellstern, Eck, Dean, Petrov, and Fiedel]{palm}
Chowdhery, A., Narang, S., Devlin, J., Bosma, M., Mishra, G., Roberts, A., Barham, P., Chung, H.~W., Sutton, C., Gehrmann, S., Schuh, P., Shi, K., Tsvyashchenko, S., Maynez, J., Rao, A., Barnes, P., Tay, Y., Shazeer, N., Prabhakaran, V., Reif, E., Du, N., Hutchinson, B., Pope, R., Bradbury, J., Austin, J., Isard, M., Gur-Ari, G., Yin, P., Duke, T., Levskaya, A., Ghemawat, S., Dev, S., Michalewski, H., Garcia, X., Misra, V., Robinson, K., Fedus, L., Zhou, D., Ippolito, D., Luan, D., Lim, H., Zoph, B., Spiridonov, A., Sepassi, R., Dohan, D., Agrawal, S., Omernick, M., Dai, A.~M., Pillai, T.~S., Pellat, M., Lewkowycz, A., Moreira, E., Child, R., Polozov, O., Lee, K., Zhou, Z., Wang, X., Saeta, B., Diaz, M., Firat, O., Catasta, M., Wei, J., Meier-Hellstern, K., Eck, D., Dean, J., Petrov, S., and Fiedel, N.
\newblock {P}a{LM}: Scaling language modeling with pathways.
\newblock Oct 2022.
\newblock \doi{10.48550/arXiv.2204.02311}.
\newblock URL \url{http://arxiv.org/abs/2204.02311}.
\newblock arXiv:2204.02311 [cs].

\bibitem[Devlin et~al.(2019)Devlin, Chang, Lee, and Toutanova]{bert}
Devlin, J., Chang, M.-W., Lee, K., and Toutanova, K.
\newblock {BERT}: Pre-training of deep bidirectional transformers for language understanding.
\newblock May 2019.
\newblock \doi{10.48550/arXiv.1810.04805}.
\newblock URL \url{http://arxiv.org/abs/1810.04805}.
\newblock arXiv:1810.04805 [cs].

\bibitem[Ding et~al.(2023)Ding, Wang, Ahmad, Ramanathan, Nallapati, Bhatia, Roth, and Xiang]{cocomic}
Ding, Y., Wang, Z., Ahmad, W.~U., Ramanathan, M.~K., Nallapati, R., Bhatia, P., Roth, D., and Xiang, B.
\newblock {C}o{C}o{MIC}: Code completion by jointly modeling in-file and cross-file context.
\newblock \penalty0 (arXiv:2212.10007), May 2023.
\newblock \doi{10.48550/arXiv.2212.10007}.
\newblock URL \url{http://arxiv.org/abs/2212.10007}.
\newblock arXiv:2212.10007 [cs].

\bibitem[Du et~al.(2022)Du, Qian, Liu, Ding, Qiu, Yang, and Tang]{glm}
Du, Z., Qian, Y., Liu, X., Ding, M., Qiu, J., Yang, Z., and Tang, J.
\newblock {GLM}: General language model pretraining with autoregressive blank infilling.
\newblock \penalty0 (arXiv:2103.10360), March 2022.
\newblock \doi{10.48550/arXiv.2103.10360}.
\newblock URL \url{http://arxiv.org/abs/2103.10360}.
\newblock arXiv:2103.10360 [cs].

\bibitem[Fried et~al.(2023)Fried, Aghajanyan, Lin, Wang, Wallace, Shi, Zhong, Yih, Zettlemoyer, and Lewis]{incoder}
Fried, D., Aghajanyan, A., Lin, J., Wang, S., Wallace, E., Shi, F., Zhong, R., Yih, W.-t., Zettlemoyer, L., and Lewis, M.
\newblock {I}n{C}oder: A generative model for code infilling and synthesis.
\newblock \penalty0 (arXiv:2204.05999), April 2023.
\newblock \doi{10.48550/arXiv.2204.05999}.
\newblock URL \url{http://arxiv.org/abs/2204.05999}.
\newblock arXiv:2204.05999 [cs].

\bibitem[Gong et~al.(2024{\natexlab{a}})Gong, Elhoushi, and Cheung]{astt5}
Gong, L., Elhoushi, M., and Cheung, A.
\newblock {AST}-{T}5: Structure-aware pretraining for code generation and understanding.
\newblock \penalty0 (arXiv:2401.03003), January 2024{\natexlab{a}}.
\newblock \doi{10.48550/arXiv.2401.03003}.
\newblock URL \url{http://arxiv.org/abs/2401.03003}.
\newblock arXiv:2401.03003 [cs].

\bibitem[Gong et~al.(2024{\natexlab{b}})Gong, Wang, and Cheung]{adelt}
Gong, L., Wang, J., and Cheung, A.
\newblock {ADELT}: Transpilation between deep learning frameworks.
\newblock \penalty0 (arXiv:2303.03593), May 2024{\natexlab{b}}.
\newblock \doi{10.48550/arXiv.2303.03593}.
\newblock URL \url{http://arxiv.org/abs/2303.03593}.
\newblock arXiv:2303.03593 [cs].

\bibitem[Gunasekar et~al.(2023)Gunasekar, Zhang, Aneja, Mendes, Del~Giorno, Gopi, Javaheripi, Kauffmann, de~Rosa, Saarikivi, Salim, Shah, Behl, Wang, Bubeck, Eldan, Kalai, Lee, and Li]{phi}
Gunasekar, S., Zhang, Y., Aneja, J., Mendes, C. C.~T., Del~Giorno, A., Gopi, S., Javaheripi, M., Kauffmann, P., de~Rosa, G., Saarikivi, O., Salim, A., Shah, S., Behl, H.~S., Wang, X., Bubeck, S., Eldan, R., Kalai, A.~T., Lee, Y.~T., and Li, Y.
\newblock Textbooks are all you need.
\newblock \penalty0 (arXiv:2306.11644), October 2023.
\newblock \doi{10.48550/arXiv.2306.11644}.
\newblock URL \url{http://arxiv.org/abs/2306.11644}.
\newblock arXiv:2306.11644 [cs].

\bibitem[Guo et~al.(2024)Guo, Zhu, Yang, Xie, Dong, Zhang, Chen, Bi, Wu, Li, Luo, Xiong, and Liang]{deepseekcoder}
Guo, D., Zhu, Q., Yang, D., Xie, Z., Dong, K., Zhang, W., Chen, G., Bi, X., Wu, Y., Li, Y.~K., Luo, F., Xiong, Y., and Liang, W.
\newblock {D}eep{S}eek-{C}oder: When the large language model meets programming -- the rise of code intelligence.
\newblock \penalty0 (arXiv:2401.14196), January 2024.
\newblock \doi{10.48550/arXiv.2401.14196}.
\newblock URL \url{http://arxiv.org/abs/2401.14196}.
\newblock arXiv:2401.14196 [cs].

\bibitem[Hendrycks et~al.(2021)Hendrycks, Basart, Kadavath, Mazeika, Arora, Guo, Burns, Puranik, He, Song, and Steinhardt]{apps}
Hendrycks, D., Basart, S., Kadavath, S., Mazeika, M., Arora, A., Guo, E., Burns, C., Puranik, S., He, H., Song, D., and Steinhardt, J.
\newblock Measuring coding challenge competence with {APPS}.
\newblock \penalty0 (arXiv:2105.09938), November 2021.
\newblock \doi{10.48550/arXiv.2105.09938}.
\newblock URL \url{http://arxiv.org/abs/2105.09938}.
\newblock arXiv:2105.09938 [cs].

\bibitem[Jain et~al.(2021)Jain, Vaidyanath, Iyer, Natarajan, Parthasarathy, Rajamani, and Sharma]{pandaseval}
Jain, N., Vaidyanath, S., Iyer, A., Natarajan, N., Parthasarathy, S., Rajamani, S., and Sharma, R.
\newblock {J}igsaw: Large language models meet program synthesis.
\newblock \penalty0 (arXiv:2112.02969), December 2021.
\newblock \doi{10.48550/arXiv.2112.02969}.
\newblock URL \url{http://arxiv.org/abs/2112.02969}.
\newblock arXiv:2112.02969 [cs].

\bibitem[Jiang et~al.(2024)Jiang, Sablayrolles, Roux, Mensch, Savary, Bamford, Chaplot, Casas, Hanna, Bressand, Lengyel, Bour, Lample, Lavaud, Saulnier, Lachaux, Stock, Subramanian, Yang, Antoniak, Scao, Gervet, Lavril, Wang, Lacroix, and Sayed]{mixtral}
Jiang, A.~Q., Sablayrolles, A., Roux, A., Mensch, A., Savary, B., Bamford, C., Chaplot, D.~S., Casas, D. d.~l., Hanna, E.~B., Bressand, F., Lengyel, G., Bour, G., Lample, G., Lavaud, L.~R., Saulnier, L., Lachaux, M.-A., Stock, P., Subramanian, S., Yang, S., Antoniak, S., Scao, T.~L., Gervet, T., Lavril, T., Wang, T., Lacroix, T., and Sayed, W.~E.
\newblock {M}ixtral of experts.
\newblock \penalty0 (arXiv:2401.04088), January 2024.
\newblock \doi{10.48550/arXiv.2401.04088}.
\newblock URL \url{http://arxiv.org/abs/2401.04088}.
\newblock arXiv:2401.04088 [cs].

\bibitem[Jimenez et~al.(2023)Jimenez, Yang, Wettig, Yao, Pei, Press, and Narasimhan]{swebench}
Jimenez, C.~E., Yang, J., Wettig, A., Yao, S., Pei, K., Press, O., and Narasimhan, K.
\newblock {SWE}-{B}ench: Can language models resolve real-world {G}ithub issues?
\newblock \penalty0 (arXiv:2310.06770), October 2023.
\newblock \doi{10.48550/arXiv.2310.06770}.
\newblock URL \url{http://arxiv.org/abs/2310.06770}.
\newblock arXiv:2310.06770 [cs].

\bibitem[Khan et~al.(2023)Khan, Bari, Do, Wang, Parvez, and Joty]{xcodeval}
Khan, M. A.~M., Bari, M.~S., Do, X.~L., Wang, W., Parvez, M.~R., and Joty, S.
\newblock x{C}ode{E}val: A large scale multilingual multitask benchmark for code understanding, generation, translation and retrieval.
\newblock \penalty0 (arXiv:2303.03004), November 2023.
\newblock \doi{10.48550/arXiv.2303.03004}.
\newblock URL \url{http://arxiv.org/abs/2303.03004}.
\newblock arXiv:2303.03004 [cs].

\bibitem[Kocetkov et~al.(2022)Kocetkov, Li, Allal, Li, Mou, Ferrandis, Jernite, Mitchell, Hughes, Wolf, Bahdanau, von Werra, and de~Vries]{thestack}
Kocetkov, D., Li, R., Allal, L.~B., Li, J., Mou, C., Ferrandis, C.~M., Jernite, Y., Mitchell, M., Hughes, S., Wolf, T., Bahdanau, D., von Werra, L., and de~Vries, H.
\newblock The {S}tack: 3 {TB} of permissively licensed source code.
\newblock \penalty0 (arXiv:2211.15533), November 2022.
\newblock \doi{10.48550/arXiv.2211.15533}.
\newblock URL \url{http://arxiv.org/abs/2211.15533}.
\newblock arXiv:2211.15533 [cs].

\bibitem[Lai et~al.(2022)Lai, Li, Wang, Zhang, Zhong, Zettlemoyer, Yih, Fried, Wang, and Yu]{ds1000}
Lai, Y., Li, C., Wang, Y., Zhang, T., Zhong, R., Zettlemoyer, L., Yih, S. W.-t., Fried, D., Wang, S., and Yu, T.
\newblock {DS}-1000: A natural and reliable benchmark for data science code generation.
\newblock \penalty0 (arXiv:2211.11501), November 2022.
\newblock \doi{10.48550/arXiv.2211.11501}.
\newblock URL \url{http://arxiv.org/abs/2211.11501}.
\newblock arXiv:2211.11501 [cs].

\bibitem[Li et~al.(2023)Li, Allal, Zi, Muennighoff, Kocetkov, Mou, Marone, Akiki, Li, Chim, Liu, Zheltonozhskii, Zhuo, Wang, Dehaene, Davaadorj, Lamy-Poirier, Monteiro, Shliazhko, Gontier, Meade, Zebaze, Yee, Umapathi, Zhu, Lipkin, Oblokulov, Wang, Murthy, Stillerman, Patel, Abulkhanov, Zocca, Dey, Zhang, Fahmy, Bhattacharyya, Yu, Singh, Luccioni, Villegas, Kunakov, Zhdanov, Romero, Lee, Timor, Ding, Schlesinger, Schoelkopf, Ebert, Dao, Mishra, Gu, Robinson, Anderson, Dolan-Gavitt, Contractor, Reddy, Fried, Bahdanau, Jernite, Ferrandis, Hughes, Wolf, Guha, von Werra, and de~Vries]{starcoder}
Li, R., Allal, L.~B., Zi, Y., Muennighoff, N., Kocetkov, D., Mou, C., Marone, M., Akiki, C., Li, J., Chim, J., Liu, Q., Zheltonozhskii, E., Zhuo, T.~Y., Wang, T., Dehaene, O., Davaadorj, M., Lamy-Poirier, J., Monteiro, J., Shliazhko, O., Gontier, N., Meade, N., Zebaze, A., Yee, M.-H., Umapathi, L.~K., Zhu, J., Lipkin, B., Oblokulov, M., Wang, Z., Murthy, R., Stillerman, J., Patel, S.~S., Abulkhanov, D., Zocca, M., Dey, M., Zhang, Z., Fahmy, N., Bhattacharyya, U., Yu, W., Singh, S., Luccioni, S., Villegas, P., Kunakov, M., Zhdanov, F., Romero, M., Lee, T., Timor, N., Ding, J., Schlesinger, C., Schoelkopf, H., Ebert, J., Dao, T., Mishra, M., Gu, A., Robinson, J., Anderson, C.~J., Dolan-Gavitt, B., Contractor, D., Reddy, S., Fried, D., Bahdanau, D., Jernite, Y., Ferrandis, C.~M., Hughes, S., Wolf, T., Guha, A., von Werra, L., and de~Vries, H.
\newblock {S}tar{C}oder: may the source be with you!
\newblock \penalty0 (arXiv:2305.06161), December 2023.
\newblock \doi{10.48550/arXiv.2305.06161}.
\newblock URL \url{http://arxiv.org/abs/2305.06161}.
\newblock arXiv:2305.06161 [cs].

\bibitem[Li et~al.(2022)Li, Choi, Chung, Kushman, Schrittwieser, Leblond, Eccles, Keeling, Gimeno, Lago, Hubert, Choy, d’Autume, Babuschkin, Chen, Huang, Welbl, Gowal, Cherepanov, Molloy, Mankowitz, Robson, Kohli, de~Freitas, Kavukcuoglu, and Vinyals]{alphacode}
Li, Y., Choi, D., Chung, J., Kushman, N., Schrittwieser, J., Leblond, R., Eccles, T., Keeling, J., Gimeno, F., Lago, A.~D., Hubert, T., Choy, P., d’Autume, C. d.~M., Babuschkin, I., Chen, X., Huang, P.-S., Welbl, J., Gowal, S., Cherepanov, A., Molloy, J., Mankowitz, D.~J., Robson, E.~S., Kohli, P., de~Freitas, N., Kavukcuoglu, K., and Vinyals, O.
\newblock Competition-level code generation with {A}lpha{C}ode.
\newblock \emph{Science}, 378\penalty0 (6624):\penalty0 1092–1097, December 2022.
\newblock ISSN 0036-8075, 1095-9203.
\newblock \doi{10.1126/science.abq1158}.
\newblock URL \url{http://arxiv.org/abs/2203.07814}.
\newblock arXiv:2203.07814 [cs].

\bibitem[Liu et~al.(2023)Liu, Xu, and McAuley]{repobench}
Liu, T., Xu, C., and McAuley, J.
\newblock {R}epo{B}ench: Benchmarking repository-level code auto-completion systems.
\newblock \penalty0 (arXiv:2306.03091), October 2023.
\newblock \doi{10.48550/arXiv.2306.03091}.
\newblock URL \url{http://arxiv.org/abs/2306.03091}.
\newblock arXiv:2306.03091 [cs].

\bibitem[Luo et~al.(2023)Luo, Xu, Zhao, Sun, Geng, Hu, Tao, Ma, Lin, and Jiang]{wizardcoder}
Luo, Z., Xu, C., Zhao, P., Sun, Q., Geng, X., Hu, W., Tao, C., Ma, J., Lin, Q., and Jiang, D.
\newblock {W}izard{C}oder: Empowering code large language models with {E}vol-{I}nstruct.
\newblock \penalty0 (arXiv:2306.08568), June 2023.
\newblock \doi{10.48550/arXiv.2306.08568}.
\newblock URL \url{http://arxiv.org/abs/2306.08568}.
\newblock arXiv:2306.08568 [cs].

\bibitem[Nijkamp et~al.(2023)Nijkamp, Pang, Hayashi, Tu, Wang, Zhou, Savarese, and Xiong]{codegen}
Nijkamp, E., Pang, B., Hayashi, H., Tu, L., Wang, H., Zhou, Y., Savarese, S., and Xiong, C.
\newblock Codegen: An open large language model for code with multi-turn program synthesis.
\newblock \penalty0 (arXiv:2203.13474), February 2023.
\newblock \doi{10.48550/arXiv.2203.13474}.
\newblock URL \url{http://arxiv.org/abs/2203.13474}.
\newblock arXiv:2203.13474 [cs].

\bibitem[OpenAI(2023)]{gpt4}
OpenAI.
\newblock {GPT}-4 technical report.
\newblock \penalty0 (arXiv:2303.08774), December 2023.
\newblock \doi{10.48550/arXiv.2303.08774}.
\newblock URL \url{http://arxiv.org/abs/2303.08774}.
\newblock arXiv:2303.08774 [cs].

\bibitem[Ouyang et~al.(2022)Ouyang, Wu, Jiang, Almeida, Wainwright, Mishkin, Zhang, Agarwal, Slama, Ray, Schulman, Hilton, Kelton, Miller, Simens, Askell, Welinder, Christiano, Leike, and Lowe]{instructgpt}
Ouyang, L., Wu, J., Jiang, X., Almeida, D., Wainwright, C.~L., Mishkin, P., Zhang, C., Agarwal, S., Slama, K., Ray, A., Schulman, J., Hilton, J., Kelton, F., Miller, L., Simens, M., Askell, A., Welinder, P., Christiano, P., Leike, J., and Lowe, R.
\newblock Training language models to follow instructions with human feedback.
\newblock \penalty0 (arXiv:2203.02155), March 2022.
\newblock \doi{10.48550/arXiv.2203.02155}.
\newblock URL \url{http://arxiv.org/abs/2203.02155}.
\newblock arXiv:2203.02155 [cs].

\bibitem[Patil et~al.(2023)Patil, Zhang, Wang, and Gonzalez]{gorilla}
Patil, S.~G., Zhang, T., Wang, X., and Gonzalez, J.~E.
\newblock {G}orilla: Large language model connected with massive apis.
\newblock \penalty0 (arXiv:2305.15334), May 2023.
\newblock \doi{10.48550/arXiv.2305.15334}.
\newblock URL \url{http://arxiv.org/abs/2305.15334}.
\newblock arXiv:2305.15334 [cs].

\bibitem[Raffel et~al.(2020)Raffel, Shazeer, Roberts, Lee, Narang, Matena, Zhou, Li, and Liu]{t5}
Raffel, C., Shazeer, N., Roberts, A., Lee, K., Narang, S., Matena, M., Zhou, Y., Li, W., and Liu, P.~J.
\newblock Exploring the limits of transfer learning with a unified text-to-text transformer.
\newblock Jul 2020.
\newblock \doi{10.48550/arXiv.1910.10683}.
\newblock URL \url{http://arxiv.org/abs/1910.10683}.
\newblock arXiv:1910.10683 [cs, stat].

\bibitem[Ren et~al.(2020)Ren, Guo, Lu, Zhou, Liu, Tang, Sundaresan, Zhou, Blanco, and Ma]{codebleu}
Ren, S., Guo, D., Lu, S., Zhou, L., Liu, S., Tang, D., Sundaresan, N., Zhou, M., Blanco, A., and Ma, S.
\newblock {C}ode{BLEU}: a method for automatic evaluation of code synthesis.
\newblock \penalty0 (arXiv:2009.10297), September 2020.
\newblock \doi{10.48550/arXiv.2009.10297}.
\newblock URL \url{http://arxiv.org/abs/2009.10297}.
\newblock arXiv:2009.10297 [cs].

\bibitem[Rozière et~al.(2023)Rozière, Gehring, Gloeckle, Sootla, Gat, Tan, Adi, Liu, Remez, Rapin, Kozhevnikov, Evtimov, Bitton, Bhatt, Ferrer, Grattafiori, Xiong, Défossez, Copet, Azhar, Touvron, Martin, Usunier, Scialom, and Synnaeve]{codellama}
Rozière, B., Gehring, J., Gloeckle, F., Sootla, S., Gat, I., Tan, X.~E., Adi, Y., Liu, J., Remez, T., Rapin, J., Kozhevnikov, A., Evtimov, I., Bitton, J., Bhatt, M., Ferrer, C.~C., Grattafiori, A., Xiong, W., Défossez, A., Copet, J., Azhar, F., Touvron, H., Martin, L., Usunier, N., Scialom, T., and Synnaeve, G.
\newblock {C}ode {L}lama: Open foundation models for code.
\newblock \penalty0 (arXiv:2308.12950), August 2023.
\newblock \doi{10.48550/arXiv.2308.12950}.
\newblock URL \url{http://arxiv.org/abs/2308.12950}.
\newblock arXiv:2308.12950 [cs].

\bibitem[Sclar et~al.(2023)Sclar, Choi, Tsvetkov, and Suhr]{prompt2}
Sclar, M., Choi, Y., Tsvetkov, Y., and Suhr, A.
\newblock Quantifying language models’ sensitivity to spurious features in prompt design or: How i learned to start worrying about prompt formatting.
\newblock \penalty0 (arXiv:2310.11324), October 2023.
\newblock \doi{10.48550/arXiv.2310.11324}.
\newblock URL \url{http://arxiv.org/abs/2310.11324}.
\newblock arXiv:2310.11324 [cs].

\bibitem[Shrivastava et~al.(2023)Shrivastava, Larochelle, and Tarlow]{repolevelprompt}
Shrivastava, D., Larochelle, H., and Tarlow, D.
\newblock Repository-level prompt generation for large language models of code.
\newblock \penalty0 (arXiv:2206.12839), June 2023.
\newblock \doi{10.48550/arXiv.2206.12839}.
\newblock URL \url{http://arxiv.org/abs/2206.12839}.
\newblock arXiv:2206.12839 [cs].

\bibitem[Team et~al.(2023)Team, Anil, Borgeaud, Wu, Alayrac, Yu, Soricut, Schalkwyk, Dai, Hauth, Millican, Silver, Petrov, Johnson, Antonoglou, Schrittwieser, Glaese, Chen, Pitler, Lillicrap, Lazaridou, Firat, Molloy, Isard, Barham, Hennigan, Lee, Viola, Reynolds, Xu, Doherty, Collins, Meyer, Rutherford, Moreira, Ayoub, Goel, Tucker, Piqueras, Krikun, Barr, Savinov, Danihelka, Roelofs, White, Andreassen, von Glehn, Yagati, Kazemi, Gonzalez, and Others]{gemini}
Team, G., Anil, R., Borgeaud, S., Wu, Y., Alayrac, J.-B., Yu, J., Soricut, R., Schalkwyk, J., Dai, A.~M., Hauth, A., Millican, K., Silver, D., Petrov, S., Johnson, M., Antonoglou, I., Schrittwieser, J., Glaese, A., Chen, J., Pitler, E., Lillicrap, T., Lazaridou, A., Firat, O., Molloy, J., Isard, M., Barham, P.~R., Hennigan, T., Lee, B., Viola, F., Reynolds, M., Xu, Y., Doherty, R., Collins, E., Meyer, C., Rutherford, E., Moreira, E., Ayoub, K., Goel, M., Tucker, G., Piqueras, E., Krikun, M., Barr, I., Savinov, N., Danihelka, I., Roelofs, B., White, A., Andreassen, A., von Glehn, T., Yagati, L., Kazemi, M., Gonzalez, L., and Others.
\newblock {G}emini: A family of highly capable multimodal models.
\newblock \penalty0 (arXiv:2312.11805), December 2023.
\newblock \doi{10.48550/arXiv.2312.11805}.
\newblock URL \url{http://arxiv.org/abs/2312.11805}.
\newblock arXiv:2312.11805 [cs].

\bibitem[Wang et~al.(2021)Wang, Wang, Joty, and Hoi]{codet5}
Wang, Y., Wang, W., Joty, S., and Hoi, S. C.~H.
\newblock {C}ode{T}5: Identifier-aware unified pre-trained encoder-decoder models for code understanding and generation.
\newblock Sep 2021.
\newblock \doi{10.48550/arXiv.2109.00859}.
\newblock URL \url{http://arxiv.org/abs/2109.00859}.
\newblock arXiv:2109.00859 [cs].

\bibitem[Wang et~al.(2023{\natexlab{a}})Wang, Le, Gotmare, Bui, Li, and Hoi]{codet5p}
Wang, Y., Le, H., Gotmare, A.~D., Bui, N. D.~Q., Li, J., and Hoi, S. C.~H.
\newblock {C}ode{T}5+: Open code large language models for code understanding and generation.
\newblock May 2023{\natexlab{a}}.
\newblock \doi{10.48550/arXiv.2305.07922}.
\newblock URL \url{http://arxiv.org/abs/2305.07922}.
\newblock arXiv:2305.07922 [cs].

\bibitem[Wang et~al.(2023{\natexlab{b}})Wang, Zhou, Fried, and Neubig]{odex}
Wang, Z., Zhou, S., Fried, D., and Neubig, G.
\newblock Execution-based evaluation for open-domain code generation.
\newblock \penalty0 (arXiv:2212.10481), May 2023{\natexlab{b}}.
\newblock \doi{10.48550/arXiv.2212.10481}.
\newblock URL \url{http://arxiv.org/abs/2212.10481}.
\newblock arXiv:2212.10481 [cs].

\bibitem[Wei et~al.(2023)Wei, Wang, Liu, Ding, and Zhang]{magicoder}
Wei, Y., Wang, Z., Liu, J., Ding, Y., and Zhang, L.
\newblock {M}agicoder: Source code is all you need.
\newblock \penalty0 (arXiv:2312.02120), December 2023.
\newblock \doi{10.48550/arXiv.2312.02120}.
\newblock URL \url{http://arxiv.org/abs/2312.02120}.
\newblock arXiv:2312.02120 [cs].

\bibitem[White et~al.(2023)White, Fu, Hays, Sandborn, Olea, Gilbert, Elnashar, Spencer-Smith, and Schmidt]{prompt1}
White, J., Fu, Q., Hays, S., Sandborn, M., Olea, C., Gilbert, H., Elnashar, A., Spencer-Smith, J., and Schmidt, D.~C.
\newblock A prompt pattern catalog to enhance prompt engineering with chatgpt.
\newblock \penalty0 (arXiv:2302.11382), February 2023.
\newblock \doi{10.48550/arXiv.2302.11382}.
\newblock URL \url{http://arxiv.org/abs/2302.11382}.
\newblock arXiv:2302.11382 [cs].

\bibitem[Xu et~al.(2022)Xu, Alon, Neubig, and Hellendoorn]{polycoder}
Xu, F.~F., Alon, U., Neubig, G., and Hellendoorn, V.~J.
\newblock A systematic evaluation of large language models of code.
\newblock \penalty0 (arXiv:2202.13169), May 2022.
\newblock \doi{10.48550/arXiv.2202.13169}.
\newblock URL \url{http://arxiv.org/abs/2202.13169}.
\newblock arXiv:2202.13169 [cs].

\bibitem[Yang et~al.(2023)Yang, Chiang, Zheng, Gonzalez, and Stoica]{datacontamination}
Yang, S., Chiang, W.-L., Zheng, L., Gonzalez, J.~E., and Stoica, I.
\newblock Rethinking benchmark and contamination for language models with rephrased samples.
\newblock \penalty0 (arXiv:2311.04850), November 2023.
\newblock \doi{10.48550/arXiv.2311.04850}.
\newblock URL \url{http://arxiv.org/abs/2311.04850}.
\newblock arXiv:2311.04850 [cs].

\bibitem[Yin et~al.(2022)Yin, Li, Xiao, Rao, Wen, Shi, Howland, Bailey, Catasta, Michalewski, Polozov, and Sutton]{arcade}
Yin, P., Li, W.-D., Xiao, K., Rao, A., Wen, Y., Shi, K., Howland, J., Bailey, P., Catasta, M., Michalewski, H., Polozov, A., and Sutton, C.
\newblock Natural language to code generation in interactive data science notebooks.
\newblock \penalty0 (arXiv:2212.09248), December 2022.
\newblock \doi{10.48550/arXiv.2212.09248}.
\newblock URL \url{http://arxiv.org/abs/2212.09248}.
\newblock arXiv:2212.09248 [cs].

\bibitem[Zhang et~al.(2023{\natexlab{a}})Zhang, Chen, Zhang, Keung, Liu, Zan, Mao, Lou, and Chen]{repocoder}
Zhang, F., Chen, B., Zhang, Y., Keung, J., Liu, J., Zan, D., Mao, Y., Lou, J.-G., and Chen, W.
\newblock {R}epo{C}oder: Repository-level code completion through iterative retrieval and generation.
\newblock \penalty0 (arXiv:2303.12570), October 2023{\natexlab{a}}.
\newblock \doi{10.48550/arXiv.2303.12570}.
\newblock URL \url{http://arxiv.org/abs/2303.12570}.
\newblock arXiv:2303.12570 [cs].

\bibitem[Zhang et~al.(2023{\natexlab{b}})Zhang, Zhang, Li, Li, Li, and Jin]{numpyeval}
Zhang, K., Zhang, H., Li, G., Li, J., Li, Z., and Jin, Z.
\newblock {T}ool{C}oder: Teach code generation models to use api search tools.
\newblock \penalty0 (arXiv:2305.04032), September 2023{\natexlab{b}}.
\newblock \doi{10.48550/arXiv.2305.04032}.
\newblock URL \url{http://arxiv.org/abs/2305.04032}.
\newblock arXiv:2305.04032 [cs].

\bibitem[Zheng et~al.(2023)Zheng, Xia, Zou, Dong, Wang, Xue, Wang, Shen, Wang, Li, Su, Yang, and Tang]{codegeex}
Zheng, Q., Xia, X., Zou, X., Dong, Y., Wang, S., Xue, Y., Wang, Z., Shen, L., Wang, A., Li, Y., Su, T., Yang, Z., and Tang, J.
\newblock {C}ode{G}ee{X}: A pre-trained model for code generation with multilingual evaluations on humaneval-x.
\newblock \penalty0 (arXiv:2303.17568), March 2023.
\newblock \doi{10.48550/arXiv.2303.17568}.
\newblock URL \url{http://arxiv.org/abs/2303.17568}.
\newblock arXiv:2303.17568 [cs].

\end{thebibliography}
